\documentclass[runningheads]{llncs}

\usepackage{graphics}
\usepackage{color}
\usepackage{fontawesome}
\usepackage{colortbl}
\usepackage{dsfont}
\usepackage{amsfonts, bm}
\usepackage{algorithm}
\usepackage[noend]{algpseudocode}
\usepackage[title]{appendix}
\usepackage{stackengine}
\usepackage{times}
\usepackage{graphicx}
\usepackage{amsmath}
\usepackage{amssymb}
\usepackage{multirow}
\usepackage{mathtools,xparse}
\usepackage{subcaption}
\usepackage{stmaryrd}
\usepackage{placeins}
\usepackage{pifont}
\usepackage{tabularx}
\usepackage{mdframed}
\usepackage{nopageno}
\usepackage{makecell}
\usepackage{arydshln}
\usepackage{booktabs}
\usepackage{enumitem}
\usepackage[skip=1ex,font=small,labelsep=period]{caption}
\usepackage{tabu}
\usepackage{etoolbox}
\usepackage{verbatim}
\usepackage{siunitx}
\usepackage{xspace}
\usepackage[table,dvipsnames]{xcolor}
\usepackage{epsfig}
\usepackage[pagebackref,breaklinks,colorlinks]{hyperref}

\let\emptyset\varnothing

\definecolor{wincolor}{rgb}{0.95, 0.2, 0.2}

\robustify\bfseries 
\newcolumntype{Y}{>{\centering\arraybackslash}X}

\usepackage[accsupp]{axessibility}  

\DeclarePairedDelimiter\abs{\lvert}{\rvert}%
\DeclarePairedDelimiter\norm{\lVert}{\rVert}%

\makeatletter
\let\oldabs\abs
\def\abs{\@ifstar{\oldabs}{\oldabs*}}
\let\oldnorm\norm
\def\norm{\@ifstar{\oldnorm}{\oldnorm*}}
\makeatother

\makeatletter
\def\BState{\State\hskip-\ALG@thistlm}
\makeatother

\algnewcommand\algorithmicforeach{\textbf{for each}}
\algdef{S}[FOR]{ForEach}[1]{\algorithmicforeach\ #1\ \algorithmicdo}

\newcommand*\ie{\emph{i.e.}\xspace}
\newcommand*\eg{\emph{e.g.}\xspace}

\newcommand{\balpha}{{\bm{\alpha}}}

\newcommand{\bmat}[1]{\begin{bmatrix}#1\end{bmatrix}}
\newcommand{\bx}{{\bm{x}}}
\newcommand{\R}{\mathbb{R}}
\newcommand{\N}{\mathbb{N}}

\newcommand{\set}[1]{\mathcal{#1}}
\newcommand{\transpose}{^\text{T}}

\begin{document}
\pagestyle{headings}
\mainmatter
\def\ECCVSubNumber{5702}  

\title{Space-Partitioning RANSAC} 

\titlerunning{Space-Partitioning RANSAC}
%
\author{Daniel Barath\inst{1} \and
Gabor Valasek\inst{2}}
\authorrunning{D\'aniel Bar\'ath and G\'abor Valasek}
%
\institute{ETH Z\"urich, Computer Vision and Geometry Group, Switzerland\\
\email{danielbela.barath@inf.ethz.ch}\\
\and
E\"otv\"os Lor\'and University, Budapest, Hungary\\
\email{valasek@inf.elte.hu}}
\maketitle

\begin{abstract}
    A new algorithm is proposed to accelerate the RANSAC model quality calculations.
    The method is based on partitioning the joint correspondence space, e.g., 2D-2D point correspondences, into a pair of regular grids.
    The grid cells are mapped by minimal sample models, estimated within RANSAC, to reject
    correspondences that are inconsistent with the model parameters early.   
    The proposed technique is general. 
    It works with arbitrary transformations even if a point is mapped to a point set, e.g., as a fundamental matrix maps to epipolar lines.
    The method is tested on thousands of image pairs from publicly available datasets on fundamental and essential matrix, homography and radially distorted homography estimation. 
    On average, the proposed space partitioning algorithm reduces the RANSAC run-time by 41\% with provably no deterioration in the accuracy.
    When combined with SPRT, the run-time drops to its 30\%.
    It can be straightforwardly plugged into any state-of-the-art RANSAC framework.
    The code is available at {https://github.com/danini/graph-cut-ransac}.
\end{abstract}

\section{Introduction}

The RANSAC (RANdom SAmple Consensus) algorithm, proposed by Fischler and Bolles~\cite{fischler1981random} in 1981, has become the most widely used robust estimator in computer vision. 
RANSAC and its variants have been successfully applied in a wide range of vision applications, such as short baseline stereo~\cite{torr1993outlier,torr1998robust}, wide baseline matching~\cite{pritchett1998wide,matas2004robust,mishkin2015mods},  performing~\cite{zuliani2005multiransac,barath2019progressive} or initializing multi-model fitting algorithms~\cite{isack2012energy,pham2014interacting}, image mosaicing~\cite{ghosh2016survey}, detection of geometric primitives~\cite{sminchisescu2005incremental}, pose-graph initialization for structure-from-motion pipelines~\cite{schoenberger2016sfm,schoenberger2016mvs}, motion segmentation~\cite{torr1993outlier}.

Briefly, RANSAC repeatedly selects random minimal subsets of the data points and fits a model to them, \eg, a 3D plane to three points, an essential matrix to five 2D point correspondences, or a rigid transformation to three 3D point correspondences.
The quality of the model is then measured, for instance, as the cardinality of its support, \ie, the number of data points closer than a manually set inlier-outlier threshold. 
Finally, the model with the highest quality, refined, \eg by least-squares fitting on all inliers, is returned.
We focus on speeding up the model quality calculation via partitioning the correspondences into pairs of $n$-dimensional cells and we select the potential inliers extremely efficiently before computing the quality of each candidate model.

\begin{figure}[t]
  \centering
      \begin{subfigure}[b]{0.325\columnwidth}
        \includegraphics[width=1.0\columnwidth]{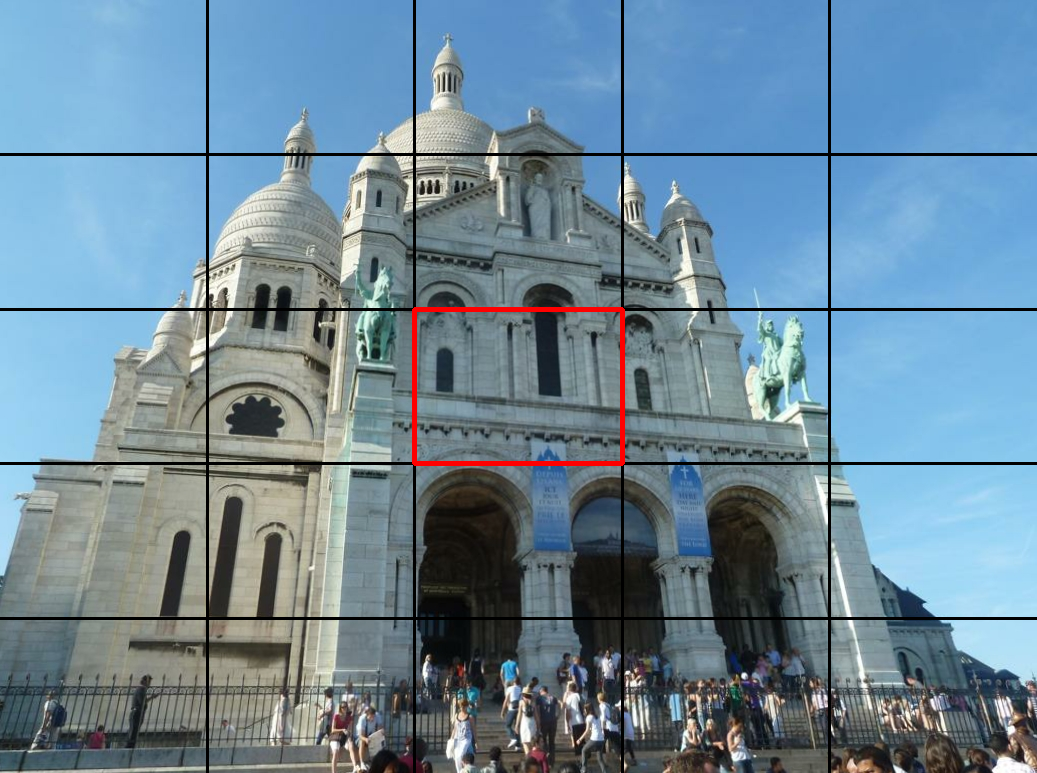}\\
        \includegraphics[width=1.0\columnwidth]{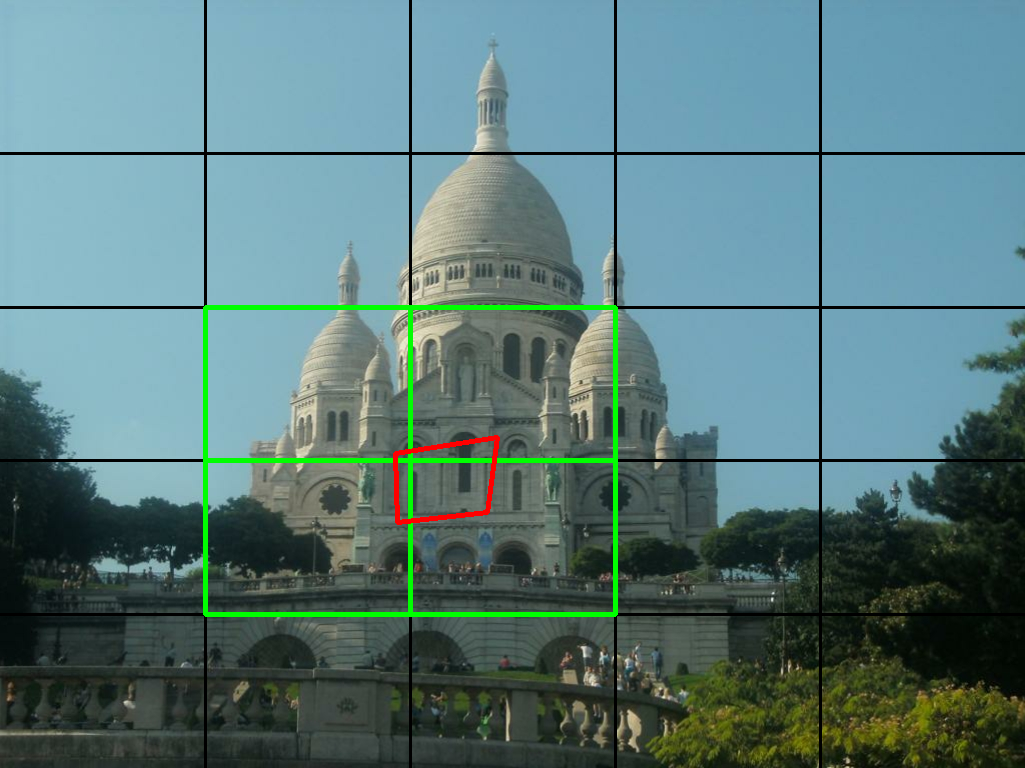}
        \caption{Homography Verification}
      \end{subfigure}
      \begin{subfigure}[b]{0.325\columnwidth}
        \includegraphics[width=1.0\columnwidth]{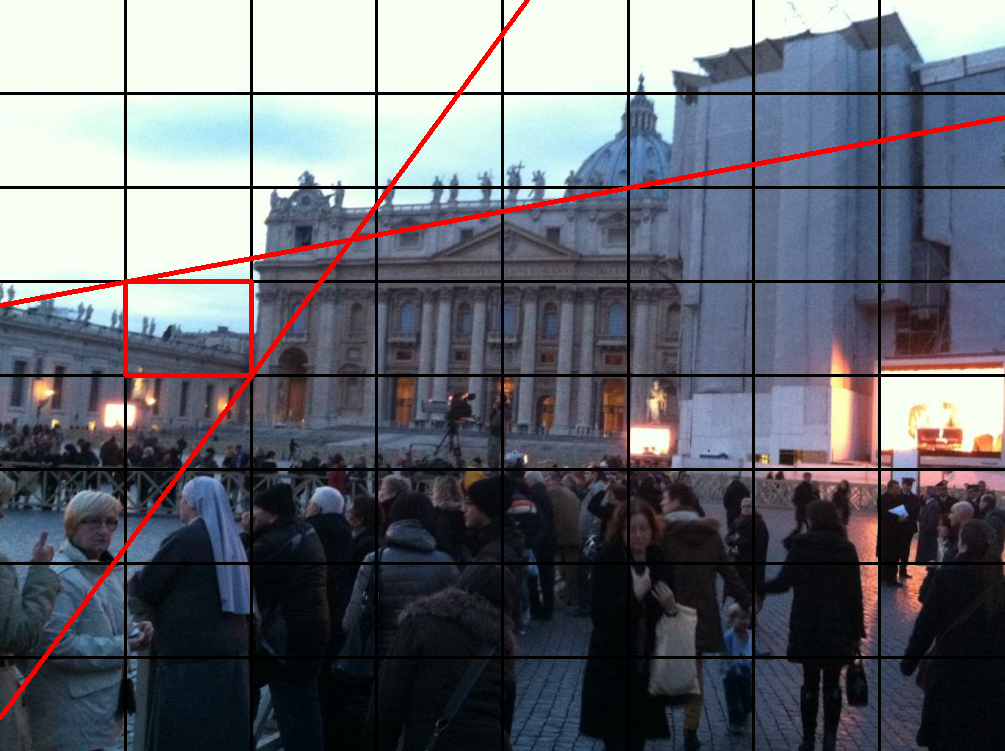}\\
        \includegraphics[width=1.0\columnwidth,trim=0 0 51mm 0,clip]{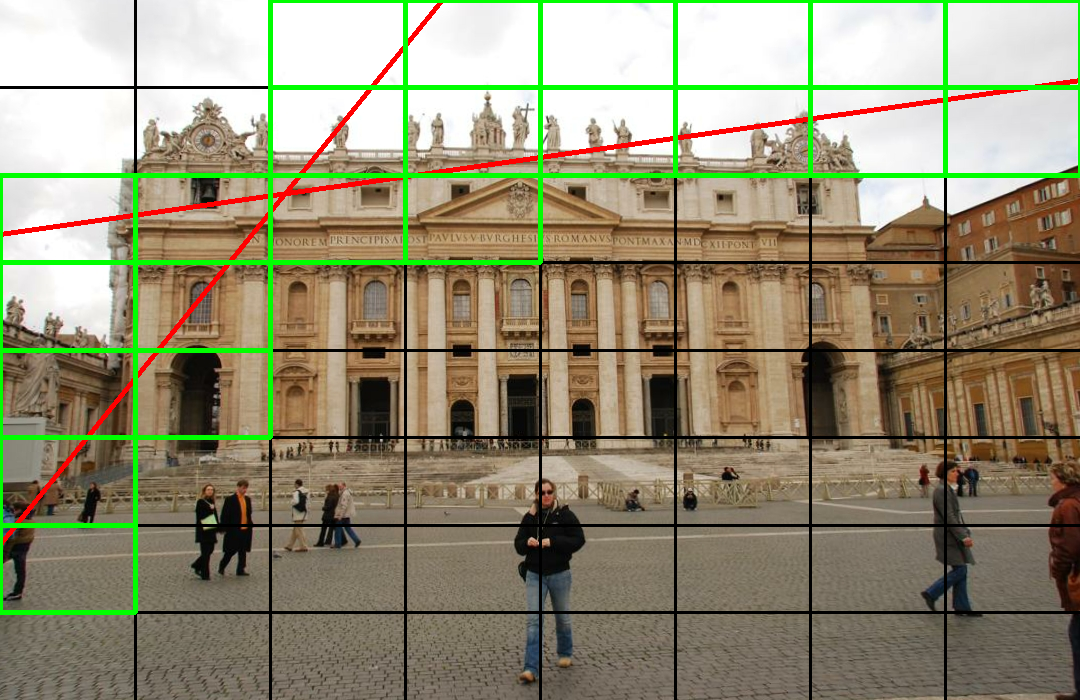}
        \caption{Epipolar Geometry Verification}
      \end{subfigure}
      \begin{subfigure}[b]{0.325\columnwidth}
        \includegraphics[width=1.0\columnwidth]{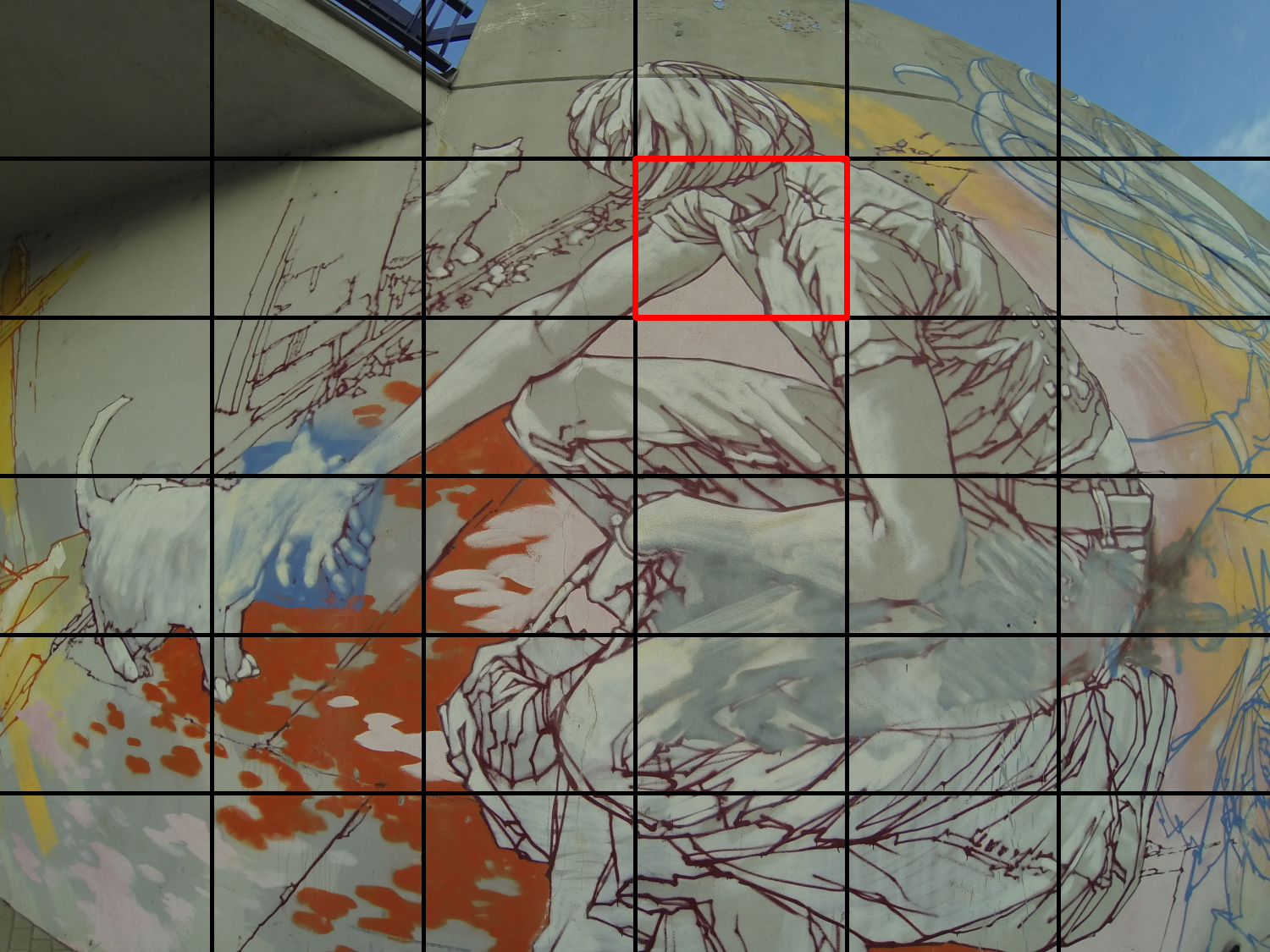}\\
        \includegraphics[width=1.0\columnwidth]{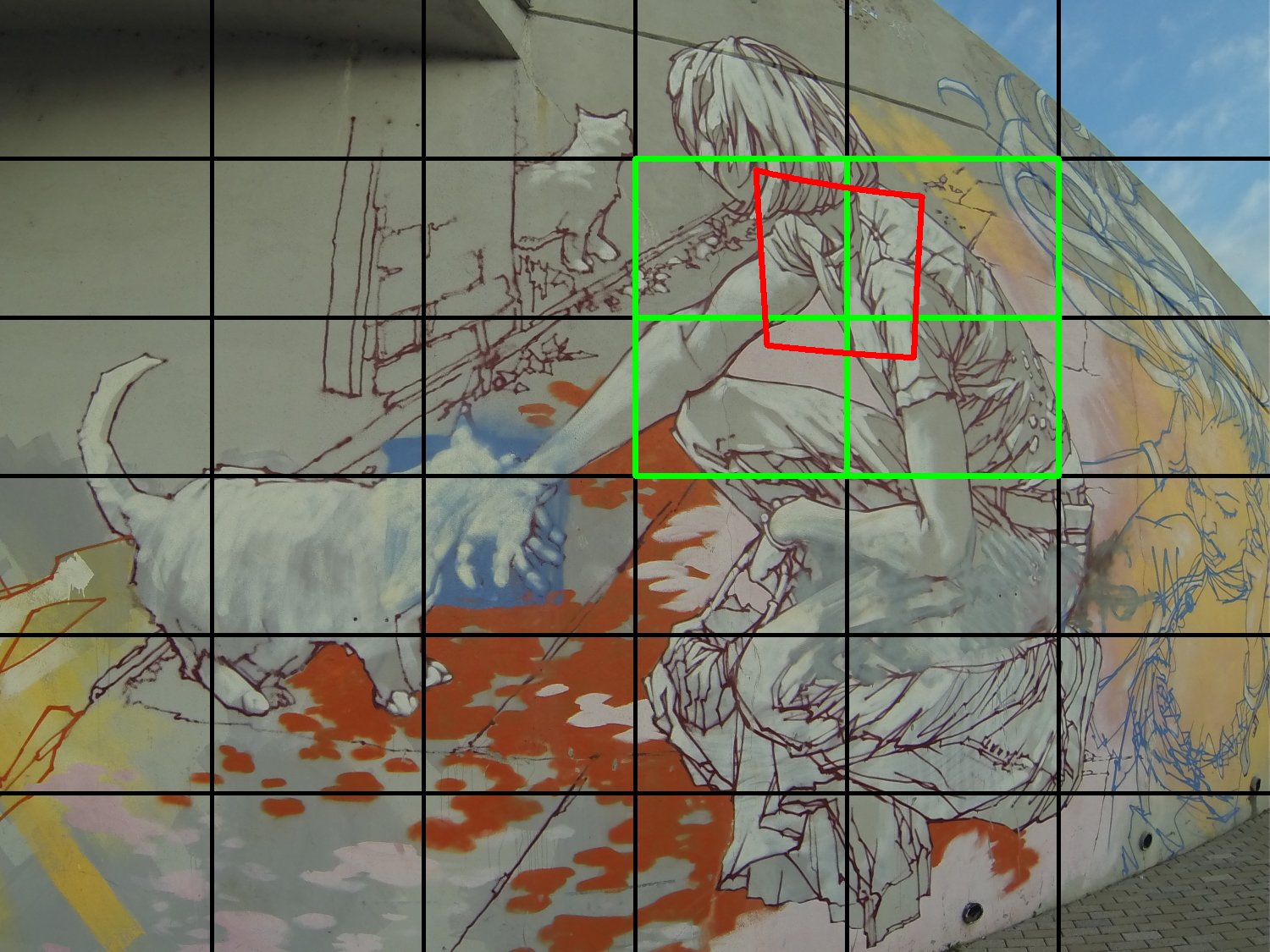}
        \caption{Radial Homography Verification}
      \end{subfigure}
   \caption{
   Examples of the proposed space partitioning-based model quality calculation.
   For a cell (red rectangle) in the top image, only those correspondences are checked where the point is in the rectangle and its corresponding pair falls inside a green rectangle in the bottom image.}
   \label{fig:culling_examples}
\end{figure}

Since the introduction of RANSAC, a number of modifications have been proposed, replacing the components of the original algorithm.
Many algorithms focus on improving the model accuracy via applying local optimization~\cite{chum2003locally,lebeda2012fixing,barath2018graph,barath2019magsac,barath2019magsacpp} that takes an initial model with reasonably high quality and improves its parameters by, \eg iterated least-squares fitting. 
To further increase accuracy, different model quality calculation techniques have been investigated by better modeling the noise in the data~\cite{torr2000mlesac,torr2002bayesian} or marginalizing over the noise scale~\cite{barath2019magsac,barath2019magsacpp}.
In the recent years, several algorithms have been proposed, including deep learning in the RANSAC procedure, \eg, as an inlier probability predictor~\cite{cne2018,dfe2018,brachmann2019ngransac,oanet2019,acne2020,clnet2021,tong2021deep}, for learning model scoring~\cite{barath2022learning} or filtering likely ill-conditioned or degenerate minimal samples early~\cite{barath2022learning,cavalli2022nefsac}.

To speed up the robust estimation procedure, several sampling algorithms have been introduced, increasing the probability of selecting a good sample early and, thus, triggering the termination criterion.
The NAPSAC~\cite{torr2002napsac} sampler assumes that inliers are spatially coherent. 
It draws samples from a hyper-sphere centered on the first, randomly selected, location-defining point. 
The GroupSAC algorithm~\cite{ni2009groupsac} assumes that inliers are often ``similar'' and, thus, data points can be separated into groups. 
PROSAC~\cite{chum2005matching} exploits an a priori predicted inlier probability rank of each point and starts the sampling with the most promising ones.
P-NAPSAC~\cite{barath2019magsacpp} merges the advantages of local and global sampling by drawing samples from gradually growing neighborhoods.

Another way of making the procedure more efficient is to avoid unnecessary calculations when computing the quality of a candidate model. 
In most of the robust estimators, the quality calculation is done for every estimated model by computing \textit{all} point-to-model residuals.
In general, this procedure is of $\mathcal{O}(NK)$ complexity, where $N$ is the number of input data points and $K$ is the number of models generated inside RANSAC. 
In the case of having thousands of input data points or a low inlier ratio, the quality calculation dominates the run-time of the robust estimation.

A number of preemptive model verification strategies have been proposed to interrupt the model quality calculation when the probability of the current model being better than the previous best falls below a threshold. 
For example, when using the $T_{D,D}$ test~\cite{chum2002randomized}, the model verification is first performed on $D$ randomly selected points (where $D \ll N$). The remaining $N - D$ ones are evaluated only if the first $D$ points are all inliers.
The test was extended by the so-called bail-out test~\cite{capel2005effective}. 
Given a model to be scored, a randomly selected subset of $D$ points is evaluated. 
If the inlier ratio within this subset is significantly smaller than the current best inlier ratio, it is unlikely that the model will yield a larger consensus set than the current best and, thus, is discarded.
In~\cite{matas2005randomized,chum2008optimal}, an optimal randomized model verification strategy was described. The test is based on Wald's theory of sequential testing~\cite{wald2004sequential}. 
Wald's Sequential Probability Ratio Test (SPRT) is a solution of a constrained optimization problem, where the user supplies acceptable probabilities for errors of the first type (rejecting a good model) and the second type (accepting a bad model) and the resulting optimal test is a trade-off between the time to decision and the errors committed. 

These methods, however, do not exploit that in computer vision the estimated model is usually an $\mathbb{R}^n \to \mathbb{R}^m$ mapping defined on geometrically interpretable data points, such as 2D-2D point correspondences.
This property allows us to partition the input points by bounding structures, \eg $n$-dimensional axis-aligned boxes (AABB), and to define the sought model as a mapping between the bounded domains. 
This is an extremely efficient way of selecting candidate inliers without calculating the point-to-model residuals of the rejected points.
The benefit is two-fold: first, significantly fewer points are needed to be tested when calculating the model quality. Second, it allows an early model rejection if the number of selected candidate inliers is lower than that of the so-far-the-best model.
Moreover, the proposed algorithm is \textit{general}.
It works for all kinds of models used in computer vision, \eg, rigid motion, homography, epipolar geometry.
It can be straightforwardly included in state-of-the-art frameworks, \eg VSAC~\cite{ivashechkin2021vsac}, and be combined with its ``bells and whistles'', \eg, the SPRT test~\cite{matas2005randomized,chum2008optimal}. 

\section{Background and Problem Formulation}

Recent robust model fitting algorithms~\cite{raguram2013usac,le2019deterministic,cai2019consensus,barath2018graph,barath2019magsac,barath2019magsacpp,ivashechkin2021vsac} spend a considerable amount of time calculating the point-to-model residuals when selecting the inliers of each verified model. 
The objective of this paper is to speed up the quality metric calculation by conservatively filtering out correspondences that are guaranteed to be outliers.

\begin{figure}[t]
  \centering
  \includegraphics[width=0.80\linewidth]{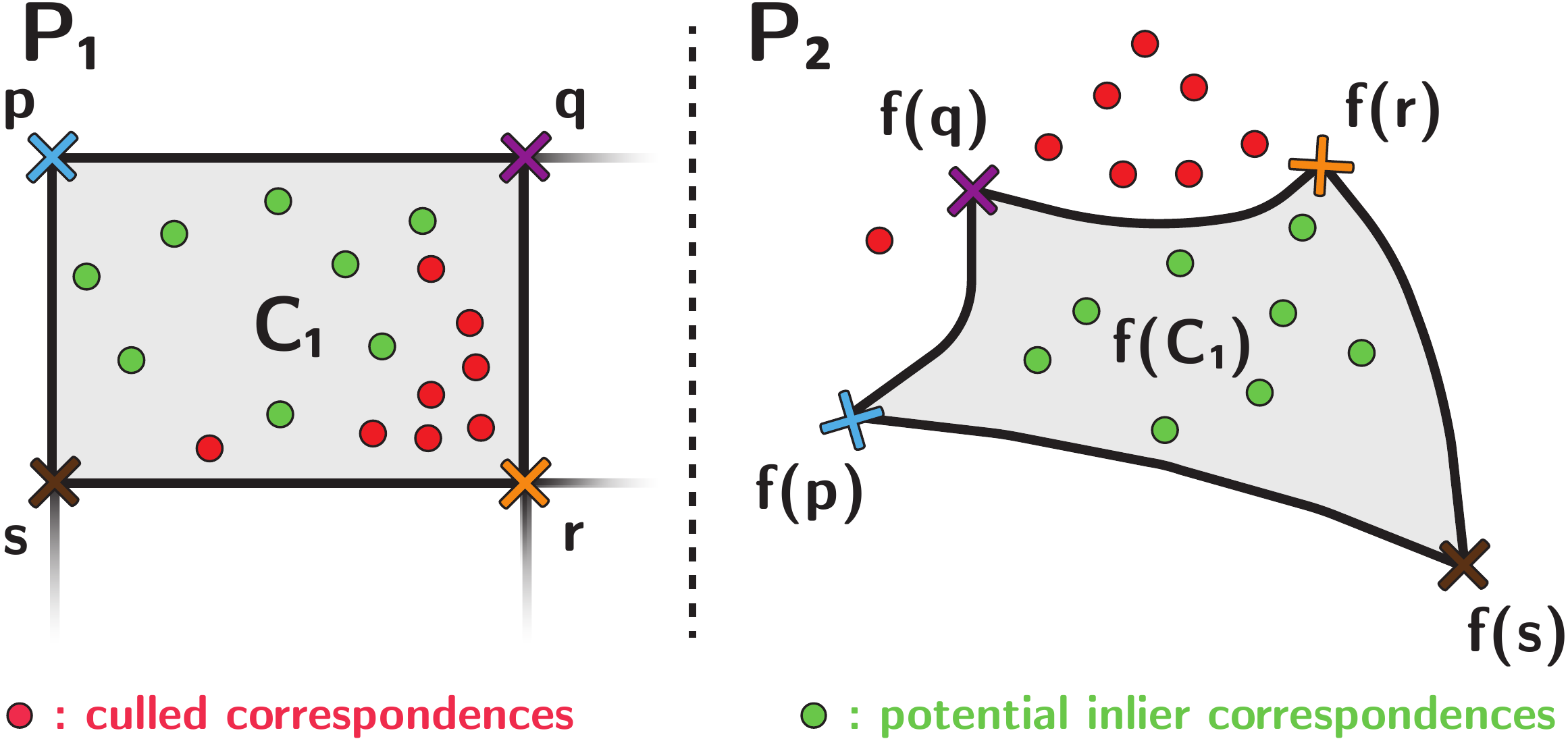}
   \caption{Cell $C_{1}$ in image plane $P_1$ and its projection $f(C_{1})$ by model $\bm f$ to $P_2$. A correspondence $(\bm p, \bm q)$ with $\bm q$ in $P_2$ falling outside $\bm f(C_{1})$ is rejected (red dots). 
   The potential inliers (green dots) of $\bm f$ are the ones where $\bm p$ falls inside $C_{1}$ and $\bm q$ in its image $\bm f(C_{1})$. }
   \label{fig:cullingByCells}
\end{figure}

For simplicity, we explain the idea through a simple example before formalizing it in a general way. 
Assume that we are given a homography, estimated from four point correspondences between two images, and the objective is to calculate the support of the model, \ie, the number of inliers.
As a preliminary step, we partition the points in each image to a regular grid, \ie, we have two grids in total. To determine the number of inliers, we process each cell of the first grid. Each cell stores a set of $(\bm p, \bm q)$ correspondences between the two images. We keep all $(\bm p, \bm q)$ pairs for which $\bm q$ is inside the projected image of the cell under the homography. More precisely, only those $(\bm p, \bm q)$ correspondences are kept as candidate inliers, where $\bm q$ is in a cell in the second image which overlaps with the projection (by the homography) of the cell in which $\bm p$ falls.
See Figures \ref{fig:culling_examples}, \ref{fig:cullingByCells} for examples.

Our formal description incorporates classes of models that map sets of points to points, so that we may address problem such as epipolar geometry estimation, where epipolar lines are mapped to points \cite{hartley2003multiple}. This comes at the expense of notational complexity but it also demonstrates the generality of our method.

\noindent\textbf{General formulation.} Let $\set{S} = \{ ( \bm p_i, \bm q_i ) \in \R^n \times \R^m \,\mid\, i=1, .., N \}$ be a set of correspondences. Points from the domain, or first image, are denoted by $\bm p_i \in \mathbb{R}^n$ and points in the range, \ie from the second image, are written as $\bm q_i \in \mathbb{R}^m$.
Set $\set{S}$ may consist of 2D-2D point pairs ($n=m=2$) found in two images and then used for estimating a homography or the epipolar geometry~\cite{hartley2003multiple}, 2D-3D correspondences ($n=2, m=3$) used to solve the perspective-\textit{n}-point problem~\cite{yuan1989general}, or 3D-3D ones ($n=m=3$) for point cloud registration~\cite{pomerleau2015review}. 
Let us assume that these correspondences stem from a $\bm g : \R^n \rightarrow \mathcal{P}( \R^m )$ ground truth mapping, where $\mathcal{P}(\R^m)$ denotes the power set of $\R^m$, \ie, the set of all possible subsets of $\R^m$. Usually, the range sets consist of a single point or posses a simple structure. For example, they form lines in the second image, such as in the case of estimating epipolar geometry where a point is mapped to an epipolar line~\cite{hartley2003multiple}.
Due to various sources of error, \eg measurement and quantization, only $\bm q_i \approx \bm g(\bm p_i)$ holds. 

Our objective is to find an $\bm f : \R^n \rightarrow \mathcal{P}( \R^m )$ \emph{model mapping} from the $(\bm p_i, \bm q_i)$ correspondences that best approximates $\bm g$ in the sense that it produces a maximal inlier set for a given $\epsilon > 0$ inlier threshold. That is, we seek the largest cardinality of the set 
\begin{equation}
    \set{I}_{\bm f} = \{ (\bm p_i, \bm q_i) \in \set{S} \mid \exists \bm q \in \bm f( \bm p_i ) : | \bm q - \bm q_i | < \epsilon \},
\end{equation}
where norm $|\cdot| : \mathbb{R}^m \to \mathbb{R}$ is some distance function defined on the points, \eg, re-projection error for homographies. 
This is the robust model fitting problem.
Note that while our model quality function maximizes the inlier number,
it is straightforward to use our proposed method with state-of-the-art functionals as well, such as the truncated quadratic loss of MSAC~\cite{torr2000mlesac} or that of MAGSAC~\cite{barath2019magsac} and MAGSAC++~\cite{barath2019magsacpp} marginalizing over a range of noise scale.

\section{Correspondence Culling}


Let $\set{P}_1 \subset \R^n$ denote the domain of the $\bm f : \R^n \rightarrow \mathcal{P}(\R^m)$ model mappings and $\set{P}_2 \subset \R^m$ the range. 
Our proposed method partitions the correspondences based on a spatial subdivision of $\set{P}_1$ into cells, for example into a regular grid. We show that the image of such regions may be conservatively bounded for a wide range of model mappings in the sense that it is guaranteed that no point of the cell may map outside of this bound.
A $(\bm p_i, \bm q_i)$ correspondence is culled if $\bm q_i \in \set{P}_2$ is outside of the bounded image of the cell containing $\bm p_i \in \set{P}_1$.
The resulting set of $n$D-$m$D point correspondences is further processed by computing the exact inlier count or model score. 

Alg.~\ref{alg:prefiltering} summarizes our approach. 
For the sake of simplicity, we will describe each step assuming that we are given 2D-2D point correspondences.
Nevertheless, the algorithm is general and, thus, also works with other data types. 
In Step~\ref{alg1:step4}, we iterate through the cells in $\set{C}_1$, \ie, the partitioning in domain $\set{P}_1$ (\ie, the first image).
Next, we calculate the bounding structure $B(\bm f(C_1), \epsilon)$ in domain $\set{P}_2$ (\ie, the second image) of the current cell $C_1$. 
In Steps~\ref{alg1:step6} and \ref{alg1:step7}, we iterate through all cells in $\set{C}_2$ and select those which intersect with $B(\bm f(C_1), \epsilon)$. 
Assuming that $\set{C}_2$ consists of axis-aligned boxes (\eg, it is a uniform grid or a quad-tree) and $B(\bm f(C_1), \epsilon)$ is an axis-align bounding box, this step simply calculates the intersection of two axis-aligned rectangles that has negligible time demand. 
In Step~\ref{alg1:step8}, we iterate through all $(\bm p_i, \bm q_i)$ correspondences where $\bm p_i$ falls inside $C_1$ and $\bm q_i$ is in $C_2$. 
Note that this step does not require checking all correspondences. 
Point correspondence $(\bm p_i, \bm q_i)$ can be considered as a 4D point $\bm s_i$ in the concatenated space $\mathbb{R}^{n+m}$, where $n = m = 2$.
Cell correspondence $(C_1, C_2)$ is basically a 4D box $C_{12}$ and, thus, the containment test degrades to checking if $\bm s_i$ falls inside $C_{12}$. 
When using hash maps and suitable hashing functions this step has $\mathcal{O}(1)$ complexity. 

\begin{algorithm}[t]
    \caption{Inlier prefiltering with conservative rejection}
    \label{alg:prefiltering}
    \begin{algorithmic}[1]
    \State \textbf{Input}: model $\bm f : \R^n \rightarrow \mathcal{P}(\R^m)$; inlier threshold  $\epsilon$; $\set{C}_1, \set{C}_2$ partitionings in images $P_1$, $P_2$.
    \State \textbf{Output}: potential inliers $\set{I}$
    \State $\set{I} \leftarrow \varnothing$
    \For {$\forall C_1 \in \set{C}_1$} \Comment{Iterate through the 2D cells in $P_1$. Section~\ref{ssec:partitioning}.}
        \label{alg1:step4}
        \State $B \leftarrow \text{Bound}( \bm f( C_1 ), \epsilon )$\Comment{Bounding structure of $\bm f(C_1)$ in $P_2$. Section~\ref{ssec:culling}.}
        \label{alg1:step5}
        \For {$\forall C_2 \in \set{C}_2$}
            \label{alg1:step6}
            \If { $C_2 \cap B \neq \emptyset$  } \Comment{Cells that intersect with $B$ in $P_2$.}
                \label{alg1:step7}
                \For {$\forall (\bm p_i, \bm q_i) \in (C_1, C_2)$} \Comment{Correspondences falling inside cells $({C}_1, {C}_2)$.}
                    \label{alg1:step8}
                    \State $\set{I} \leftarrow \set{I} \cup \{ (\bm p_i, \bm q_i) \}$
                \EndFor
            \EndIf
        \EndFor
    \EndFor
    \end{algorithmic}
\end{algorithm}

\subsection{Correspondence Partitioning} \label{ssec:partitioning}


Let us partition $\set{P}_1$ into a disjoint set of $C_j \in \set{C}$ cells ($j \in \mathbb{N}_{>0}$) and store the correspondences based on these, \ie let $\set{S}_j = \{ ( \bm p_i, \bm q_i ) \in \set{S} \mid \bm p_i \in C_j \}$.  
Set $\set{S}_j$ consists of the correspondences where the first point falls inside cell $C_j$ in $\set{P}_1$.
In our particular case, we use a regular grid of equally sized axis-aligned rectangular cells for the subdivision. 
Partitioning is a pre-processing step with $\mathcal{O}(N)$ complexity and it only needs to be computed once, for example, upon reading the point correspondences. Decomposing $\set{S}$ into the $\set{S}_{j}$ partitions is not strictly necessary, however, it is an important component of efficient culling.

Note that the proposed method can be also used with more advanced space partitioning structures, such as quad-trees. 
We, however, empirically found that the required computational overhead is too large for the typical computer vision problems consisting of, at most, a few thousands of data points. 
This overhead stems from the tree construction and the increased number of cells to be projected by the proposed algorithm. 


\subsection{Culling by Images of Cells} \label{ssec:culling}

Suppose that we are given a model $\bm f : \mathcal{P}_1 \rightarrow \mathcal{P}_2$ estimated, for example, from a minimal sample.
The image of a $C_j \subset \mathcal{P}_1$ cell depends on the current model $\bm f$ and cannot be pre-computed. Consequently, we have to devise efficient means to compute or estimate this image, as it is calculated at every RANSAC iteration.

Depending on the algebraic properties of the model mapping, there are two cases to consider. If $\bm f$ is invertible, we cull $(\bm p_i, \bm q_i), \bm p_i \in C_j$ if $\bm f^{-1}$ does not map $\bm q_i$ into $C_j$, that is, if $\bm f^{-1}(\bm q_i) \not\in C_j$. If $\bm f$ is not invertible, we bound the image of the $C_j$ cell under $\bm f$ with some $B \subset \mathcal{P}_2$ and reject $(\bm p_i, \bm q_i)$ if $\bm q_i \not\in B$ holds. This is further decomposed into two cases: $B$ may be either an approximative or a conservative bound. 

\noindent\textbf{Invertible Case.}
Let us define the image of cell $C_j$ as the image of all points in $C_j$, \ie
\begin{equation}
    \bm f(C_j) = \{ \bm y \in \R^m \mid \exists \bm x \in C_j : \bm y \in \bm f( \bm x ) \}.
\end{equation}
%
Note that $\bm x$ and $\bm y$ denote arbitrary points in $\mathbb{R}^n$ and $\mathbb{R}^m$, respectively, and not necessarily the input correspondences. 
%

Let us assume that there exists an $\bm f^{-1}$ inverse of the model map. The $\bm f^{-1}(\bm q_i) \in C_j$ containment is trivially resolved if we have a regular grid in $\mathcal{P}_1$. 
In the general case, let us assume that $C_j$ is written as the intersection of a finite number of 
\begin{equation}
    Q_{kj} = \{ \bm x \in \set{P}_1 \mid q_{kj}(\bm x) \leq 0 \}
\end{equation}
volumes, \ie $C_j = \cap_k Q_{kj}$, where $q_{kj} : \set{P}_1 \rightarrow \R$ is the implicit representation of the $j$-th boundary volume. 
This intersection contains all points inside and on the boundary of $C_j$.
For example, a 2D rectangular cell may be written as the intersection of four half-planes of the form $a_{k} x + b_{k} y + c_{k} \leq 0, k \in \{1,2,3,4\}$.
Recalling that the intersection of implicitly defined volumes may be written as a maximum operation \cite{Ricci:CSG}, a $(\bm x, \bm y) \in \set{S}_j$ correspondence is culled, if and only if 
%
    $\max\{ q_{kj}( \bm f^{-1}(\bm y) \}_{k} > 0$. 
%
Thus, the containment test in the transformed space $\set{P}_2$ is reduced to a test in $\set{P}_1$. 

Although closed-form inversion is a strict restriction, all non-singular projective transformations possess this property. 
While no speed-up is achieved by transforming all points inside a cell in this way, the definitions will later be important to efficiently cull against AABBs of the cell boundaries.


\noindent\textbf{General Case.}
%
If no closed-form inverse is available for $\bm f$, we may bound the image of cell $C_j$ by using polynomial approximations of the transformed boundaries. 
We propose to use Lagrange interpolation at Chebyshev nodes to obtain these approximations. Once these polynomial approximations are obtained, we convert them to Bernstein basis and use the resulting B\'ezier control points to form an AABB bound of the cell image. This bounding characteristic follows from the fact that B\'ezier curves possess the convex hull property \cite{Farin:CAGD}, that is, the entire image of the cell under the approximations should lie within the convex hull of the control points, and in turn, within their AABB. 

In terms of specifics to achieve the above, let us assume that the boundaries of the transformed cell, $\bm f(C_j)$, are the images of the cell boundaries in $C_j$ as $\bm f( \{ \bm x \in \set{P}_1 \mid q_{kj}(\bm x) = 0 \} )$. This merely simplifies the identification of the boundary curves of $\bm f(C_j)$.
%
%
%
This is straightforward for models that map points to points as the polynomial approximations are traditional two or three dimensional B\'ezier curves. 
Models that map points to sets of points, \eg epipolar geometry, can be embedded into this framework by assuming that there is a simple set of basis functions in which the image sets may be represented. 
For example, let us assume that the image sets are one parameter families and there exists a function basis that spans all sets. 
Formally, this means that, for all $\bm x \in \set{P}_1$, we assume the existence of a $\bm p_{\bm x} : \R \rightarrow \set{P}_2$ parametric mapping such that 
\begin{equation}
    \bm f( \bm x ) = \{ \bm y \in \mathcal{P}_2 \mid \exists t \in \R : \bm y = \bm p_{\bm x}(t) \} ~.
\end{equation}
Let $\{ e_i(t) : \R \rightarrow \R \}_{i=1}^k$ denote the basis that spans the images, \ie,  $\forall \bm x \in \set{P}_1: \forall i \in \{ 1, \dots, k \}: \exists \bm a_{\bm x,i} \in \set{P}_2:$ 
\begin{equation}
    \bm p_{\bm x}(t) = \sum_{i=1}^k a_{\bm x,i} e_i(t) ~.
\end{equation}
For example, lines in $\mathcal{P}_2$ may be represented in the $\{1, t\}$ basis as $\bm p_{\bm x}(t) = \bm p_{\bm x, 0} + t \bm v_{\bm x}$.
The polynomial approximation of the boundary is reduced to the construction of a higher dimensional B\'ezier curve that maps from $\R$ to $\R^{m k}$, \ie the B\'ezier curve is used to approximate the $a_{\bm x, i}$ coefficients. As shown in the example of epipolar geometry estimation, this can be made simpler for a particular problem. 

\noindent\textbf{Conservative Bounds.}
The polynomial approximations shown above can be made conservative by applying the Lagrange interpolation error term to them. 

If an arbitrary boundary curve of cell $C_j$ is parametrized over an interval $[a,b]$ by some $\bm q_{kj} : [a,b] \rightarrow \set{P}_1 ~ (k \in \{1,2,3,4\})$ mapping, then interpolating $\bm f \circ \bm q$ at $k$ Chebyshev nodes $\bm f( \bm q_{kj}( \bm x_l )  )$, where
\begin{equation}
     \bm x_l = \frac{a+b}{2} + \frac{b-a}{2} \cos{\frac{2l-1}{2k}\pi} , l = 1, \dots, k
\end{equation}
yields a polynomial that is within 
\begin{equation}
    \frac{1}{2^{k-1}k!} \left( \frac{b-a}{2} \right)^k \cdot \max_{\xi \in [a,b]} \lVert (\bm f \circ \bm q)^{(k+1)}(\xi) \rVert_{\infty}
\end{equation}
of a $(k+1)$ times continuously differentiable $\bm f \circ \bm q$ mapping \cite{books/daglib/0030811}. Thus, offsetting the AABB of the B\'ezier control points by this value ensures that no points in $C_j$ may map outside the AABB bounding the image $\bm f(C_j)$ of cell $C_j$. 
Moreover, to account for the inlier-outlier threshold of the robust estimation procedure set, the AABB should additionally be offsetted by the threshold.

The simplest case is when the cells are axis-aligned rectangles and the $\bm q_{kj}$ boundaries are linear interpolations between the vertices of the edges.
Algorithm \ref{alg:bound} summarizes the construction of conservative bounds to the images in such a configuration. This has to be run once for each cell. For each edge of the cell, it requires $k$ evaluations of $\bm f$. Steps \ref{step:interpolate} and $\ref{step:BernsteinConversion}$ can be carried out simultaneously by a multiplication of a $(k-1) \times (k+1)$ matrix with a $(k+1) \times 2$ matrix. Step \ref{step:derivativeBound} depends on the candidate mapping family. In the case when no closed-form solutions are available for the bound, one has to estimate it by numerical means. The bound on the image can be made tighter by using the second grid on $\set{P}_2$ and select the cells that intersect the convex hull of the control points. 

Note that when using regular grids, the proposed algorithm can be significantly sped up by projecting the boundaries shared by multiple cells only once.

\begin{algorithm}[t]
    \caption{$\text{Bound}( \bm f( C_j ), \epsilon )$ with AABB}
    \label{alg:bound}
    \begin{algorithmic}[1]
        \Statex \textbf{Input}: 
            current candidate mapping $\bm f : \R^2 \rightarrow \mathcal{P}(\R^2)$; 
        \Statex \phantom{\textbf{Input}:} approximation degree $k$; $\ C_j$ cell; $\epsilon > 0$
        \State $x_{min, j}, y_{min, j} \leftarrow +\infty$
        \State $x_{max, j}, y_{max, j} \leftarrow -\infty$
        \For { $\forall e$ edge of $C_j$ }
            \State $\bm a, \bm b \leftarrow$ the endpoints of edge $e$
            \State $\{\bm x_l \leftarrow \bm a + t_l(\bm b - \bm a) \}_{l=1}^k$; $t_l$ are Chebyshev nodes
            \State $\bm c(t) \leftarrow$ polynomial interpolating $\{ \bm f( \bm x_l ) \}_{l=1}^k$ \label{step:interpolate}
            \State $\bm b_0, \dots, \bm b_{k+1} \leftarrow$ B\'ezier control points of $\bm c(t)$ \label{step:BernsteinConversion} 
            \State $x_{min, j} \leftarrow \min \left\{ x_{min, j}, \min_{i=0}^{k+1} \{ [1,0] \cdot \bm b_i \} \right\}$
            \State $y_{min, j} \leftarrow \min \left\{ y_{min, j}, \min_{i=0}^{k+1} \{ [0,1] \cdot \bm b_i \} \right\}$
            \State $x_{max, j} \leftarrow \max \left\{ x_{min, j}, \max_{i=0}^{k+1} \{ [1,0] \cdot \bm b_i \} \right\}$
            \State $y_{max, j} \leftarrow \max \left\{ y_{min, j}, \max_{i=0}^{k+1} \{ [0,1] \cdot \bm b_i \} \right\}$
        \EndFor
        \State $M \leftarrow $ bound on $\lVert \bm f^{(k+1)}(\bm a + t( \bm b - \bm a )) \rVert_{\infty}$, $t \in [0,1]$ \label{step:derivativeBound}
        \State $C \leftarrow \max_{\bm x} \Pi_{j=0}^k |x - t_j| = \frac{1}{2^{2k-1}k!}$
        \State $\delta \leftarrow \epsilon + \lVert \bm b - \bm a \rVert_{\infty} \cdot M + C$
        \State \textbf{return} $ \left( \bmat{ x_{min, j} - \delta \\ y_{min, j}- \delta}, \bmat{ x_{max, j} + \delta \\ y_{max, j} + \delta } \right)$
    \end{algorithmic}
\end{algorithm}

\subsection{Early Model Rejection}

The proposed approach provides all cell correspondences that might contain inliers of the currently estimated model.
Besides being extremely useful for the verification, it also helps in rejecting models early without calculating their quality. 
The total number of data points stored in the selected cells is basically an upper bound on the inlier number.
In the case this bound does not exceed the inlier number of the previous so-far-the-best model, the current model can be immediately rejected as it will not have more inliers than the best model so far.
Therefore, a model $\theta$ is rejected if $\epsilon_r |\mathcal{I}^*| > |\mathcal{I}_\theta|$, where $\mathcal{I}^*$ and $\mathcal{I}_\theta$ are, respectively, the inlier sets of the so-far-the-best and the currently tested models. Parameter $\epsilon_r = 1$ provably leads to no accuracy change. However, in practice, $\epsilon_r$ can be set marginally higher as we will show in the experiments.

Note that this approach works for all quality functions where the model quality is calculated from points closer than a manually set threshold, \eg, as in RANSAC, MSAC or even in MAGSAC++ which uses a maximum threshold.

\section{Model Estimation Problems}

We show how the described general algorithm can be used in computer vision tasks. 

\subsection{Homography Estimation}

Given homography $\textbf{H} \in \mathbb{R}^{3\times3}$, the implied relationship of the points in the two images is written as $\alpha \textbf{H} \bm{x}_1 = \bm{x}_2$, where $\alpha$ is a scaling that is inverse proportional to the homogeneous coordinate. The implied mapping is as 
\begin{equation*}
    \bm{f}_{\text{hom}}(\bm{x})  = \begin{bmatrix}
        f_u(\bm{x}) \\ f_v(\bm{x})
    \end{bmatrix} =
    \begin{bmatrix}
        \bm{h}_1\transpose \bm{x} / \bm{h}_3\transpose \bm{x} \\
        \bm{h}_2\transpose \bm{x} / \bm{h}_3\transpose \bm{x} \\
    \end{bmatrix},
\end{equation*}
where $\bm{h}_1, \bm{h}_2, \bm{h}_3 \in \mathbb{R}^3$ are the rows of $\textbf{H}$, and $f_u(\bm{x})$,  $f_v(\bm{x})$ are coordinate functions.
Since $\textbf{H}$ maps lines to lines and we use a regular grid, the culling algorithm can be simplified to using the AABB of the cell corners projected by $\bm{f}_{\text{hom}}$. This is a degree-one polynomial approximation that is also exact.


\subsection{Epipolar Geometry Estimation}

Given fundamental matrix $\textbf{F}$, the implied relationship of the points in the two images is written as $\bm{x}_2\transpose \textbf{F} \bm{x}_1 = 0$, 
meaning that point $\bm{x}_2$ in the second image must fall on the corresponding epipolar line $\textbf{l}_2 = \begin{bmatrix} a_2 & b_2 & c_2 \end{bmatrix}\transpose = \textbf{F} \bm{x}_1$.
To bound the problem, we can assume that our model maps to the angle of the epipolar line as follows:
\begin{equation*}
    \bm{f}_{\text{epi}}(\bm{x}) = \tan^{-1} \frac{b_2(\bm{x})}{a_2(\bm{x})}, \quad \bm{f}_{\text{epi}}^{-1}(\bm{x}) = \tan^{-1} \frac{b_1(\bm{x})}{a_1(\bm{x})},
\end{equation*}
where $\textbf{l}_i(\bm{x}) = \begin{bmatrix} a_i(\bm{x}) & b_i(\bm{x}) & c_i(\bm{x}) \end{bmatrix}$ is the epipolar line implied by $\bm{x}$ in the $i$th image.
Given cell $C_j$ containing the epipolar angles, it can be straightforwardly seen that
\begin{eqnarray*}
    \bm{f}_{\text{epi}}(C_j) : [\min \{\alpha \in C_j \}, \max \{\alpha \in C_j \}] \to  
    \left[ \min \{\alpha \in \bm{f}_{\text{epi}}(C_j) \}, \max \{ \alpha \in \bm{f}_{\text{epi}}(C_j) \} \right].
\end{eqnarray*}
Due to the convexity of the problem and the used regular grid, it can be decomposed into two sub-problems.
First, the $\min$ and $\max$ operations are performed only on the intersections of the boundaries, \ie, the cell corners.
Therefore, a cell is not culled if at least one epipolar angle implied by its corners fall inside interval $[\min \{\alpha \in \bm{f}_{\text{epi}}(C_j) \}, \max \{ \alpha \in \bm{f}_{\text{epi}}(C_j) \}]$.
Second, a cell is not culled if one of its boundary lines intersects with the epipolar lines implied by angles $\min \{\alpha \in \bm{f}_{\text{epi}}(C_j) \}$ and $\max \{\alpha \in \bm{f}_{\text{epi}}(C_j) \}$.
This is a more efficient formulation of the problem than considering epipolar lines as point sets. 

\subsection{Radial Homography Estimation}

We use the one-parameter division model~\cite{fitzgibbon2001simultaneous}, for modeling radial distortion, is of form
%
    $\bm{g}(\bm{x}, \lambda) = [ u, v, 1 + \lambda (x^2 + y^2)]\transpose$,
%
where $\bm x = [u, \, v]\transpose$ is an image point.
Given homography $\textbf{H} = \left[ \bm{h}_1, \bm{h}_2, \bm{h}_3 \right]\transpose$ with rows $\bm{h}_1, \bm{h}_2, \bm{h}_3 \in \mathbb{R}^3$, the points in the images are related as 
%
    $\lambda_2 \bm{g}(\bm{x}_2, \lambda_2) = \textbf{H} \bm{g}(\bm{x}_1, \lambda_1)$,
%
where $\bm{x}_i$ and $\lambda_i$ are, respectively, the point and the distortion parameter in the $i$th image, $i \in [1, 2]$. 
This implies the following mapping function
%
    $\bm{f}_{\text{rad}}(\bm{x}) = [
        f_u(\bm{x}), f_v(\bm{x})]\transpose$,
%
where $f_u, f_v : \mathbb{R} \to \mathbb{R}$ are coordinate functions as follows:
\begin{equation}
    f_u(\bm{x}) = \frac{\bm{h}_1\transpose \bm{g}(\bm{x})}{\bm{h}_3\transpose \bm{g}(\bm{x})}, \quad f_v(\bm{x}) = \frac{\bm{h}_2\transpose \bm{g}(\bm{x})}{\bm{h}_3\transpose \bm{g}(\bm{x})}.
    \label{eq:radial_mapping}
\end{equation}
The conservative bounds described in the previous sections are calculated from \eqref{eq:radial_mapping}.

\section{Experiments}

We compare the proposed space partitioning-based model verification technique to the traditional one and to the SPRT~\cite{chum2008optimal} algorithm, \ie, the state-of-the-art preemptive model verification technique.
Besides comparing to SPRT, we also show that combining the two methods is highly beneficial.
For robust estimation, we use LO-RANSAC~\cite{chum2003locally} with PROSAC~\cite{chum2005matching} sampling.
%
We do not report model accuracy since the proposed technique leads to exactly the same number of inliers as verifying all points.

\begin{figure*}[t]
  \centering
      \begin{subfigure}[b]{0.32\linewidth}
     	 	\centering
  	        \includegraphics[width=1.0\columnwidth,trim=0 0 0cm 0cm, clip]{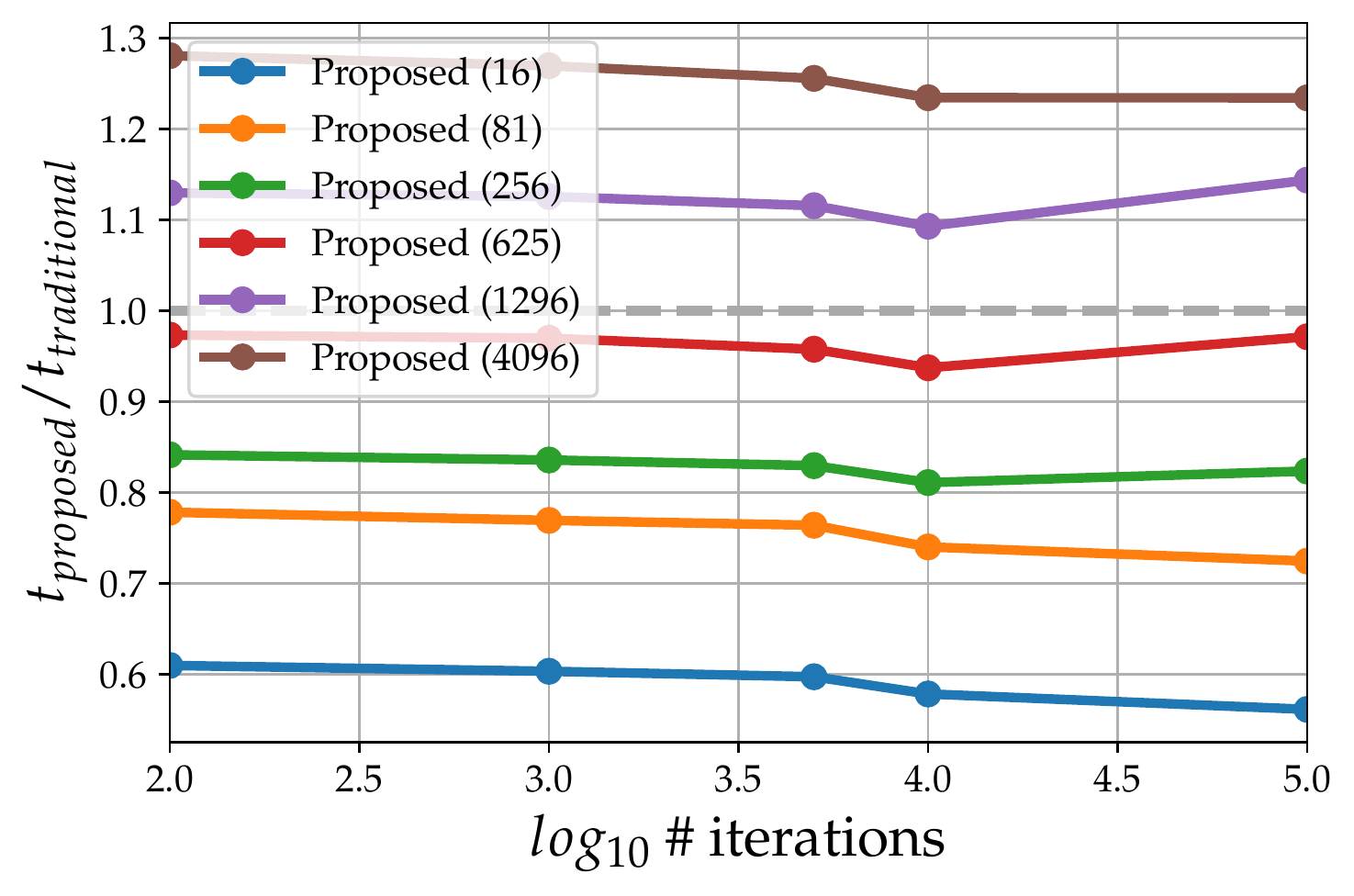}\\
  	        \includegraphics[width=1.0\columnwidth,trim=0 0 0cm 0cm, clip]{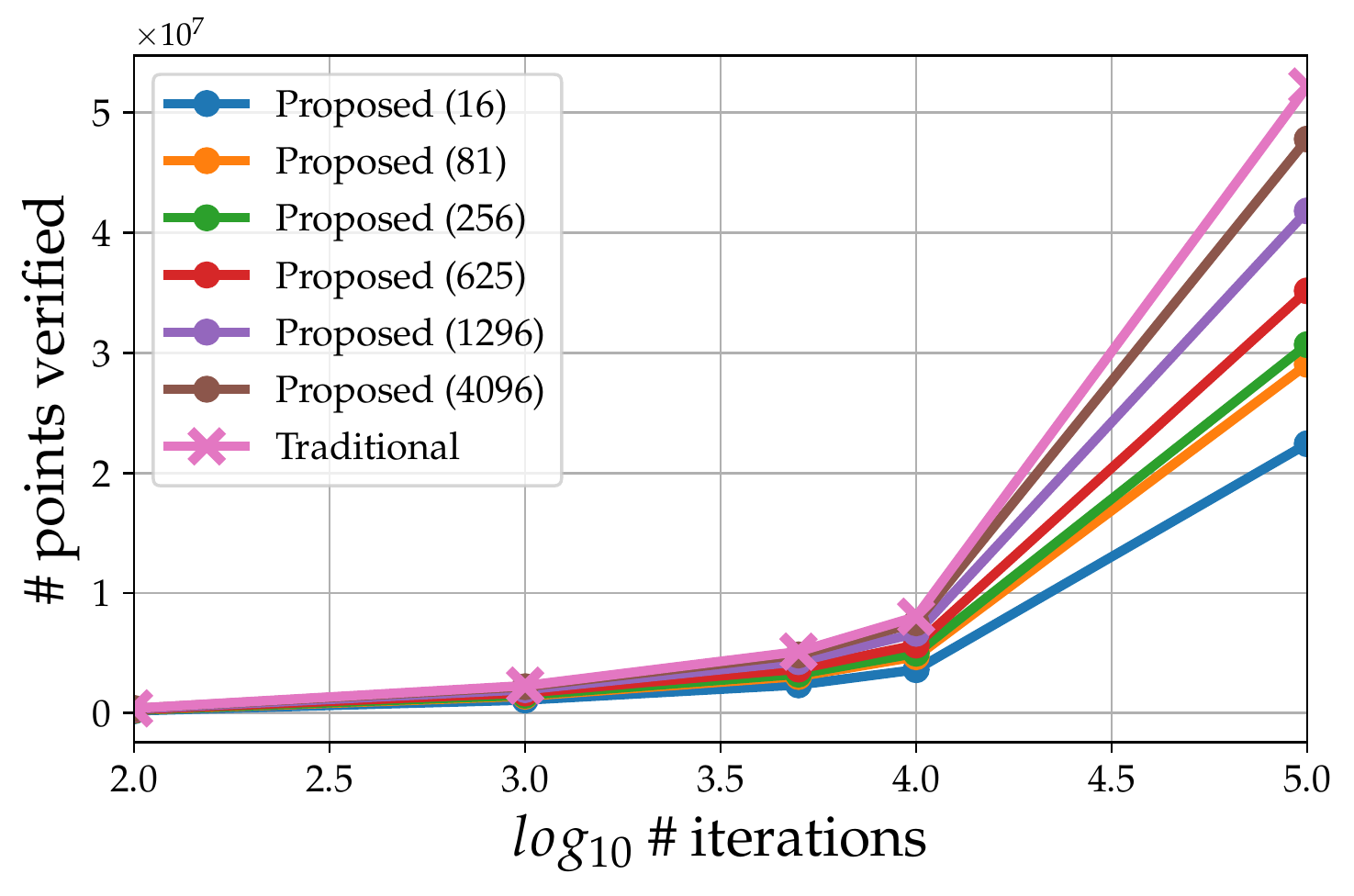}\\
  	        \includegraphics[width=1.0\columnwidth,trim=0 0 0cm 0cm, clip]{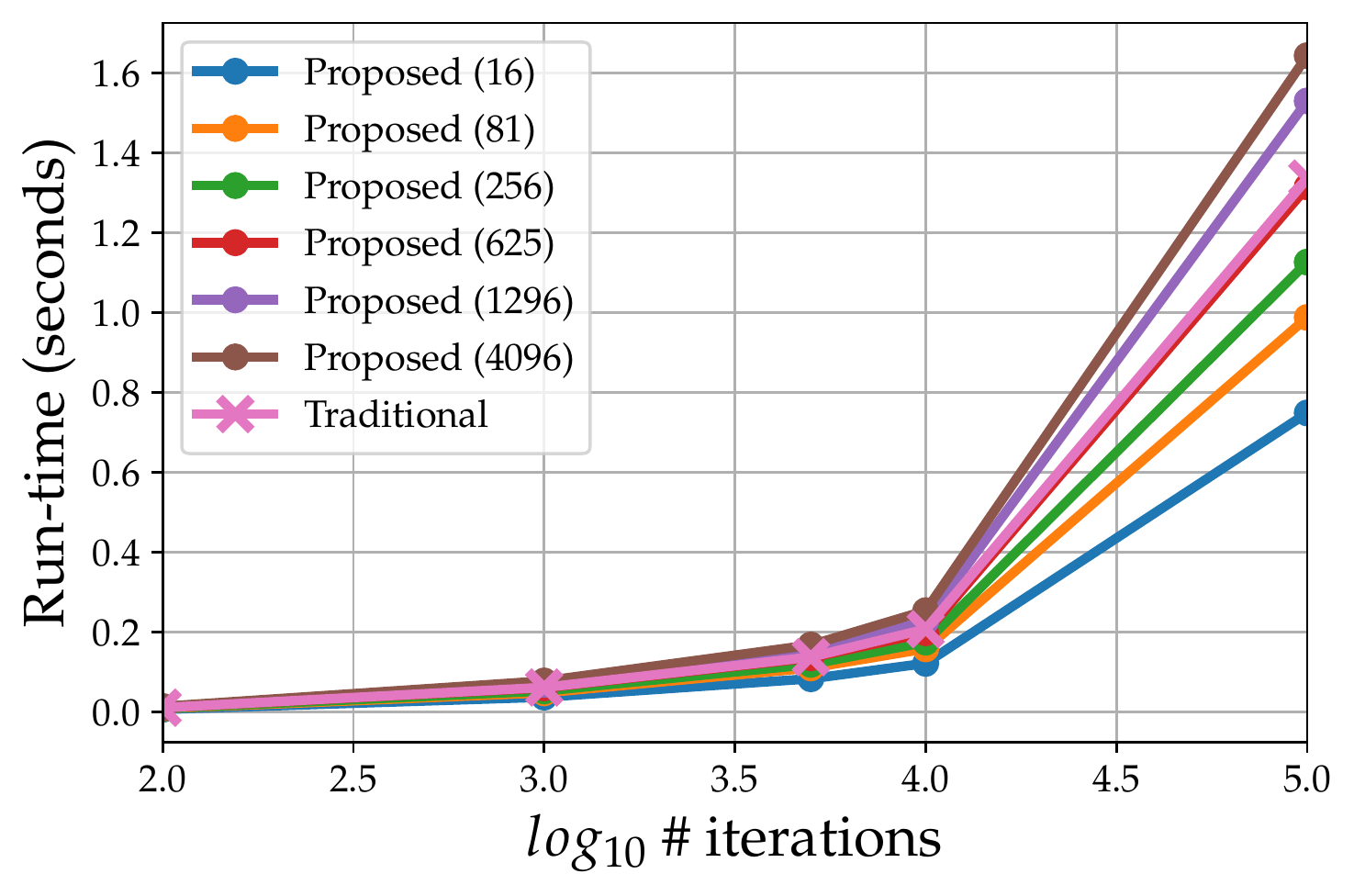}
      		\caption{Fundamental matrix estimation}
            \label{fig:resultsF}
      \end{subfigure}
      \begin{subfigure}[b]{0.32\linewidth}
     	 	\centering
  	        \includegraphics[width=1.0\columnwidth,trim=0 0 0cm 0cm, clip]{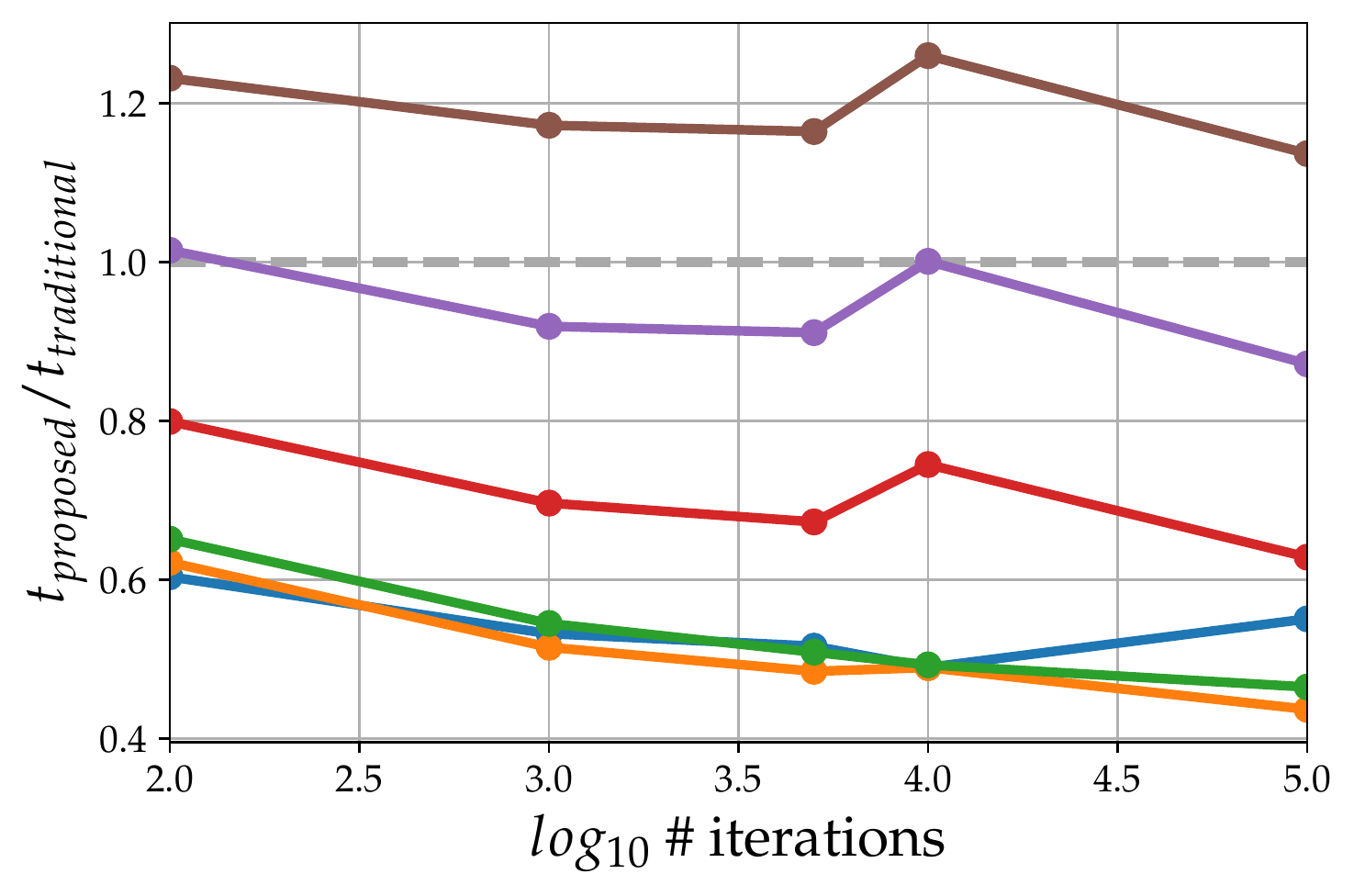}\\
  	        \includegraphics[width=1.0\columnwidth,trim=0 0 0cm 0cm, clip]{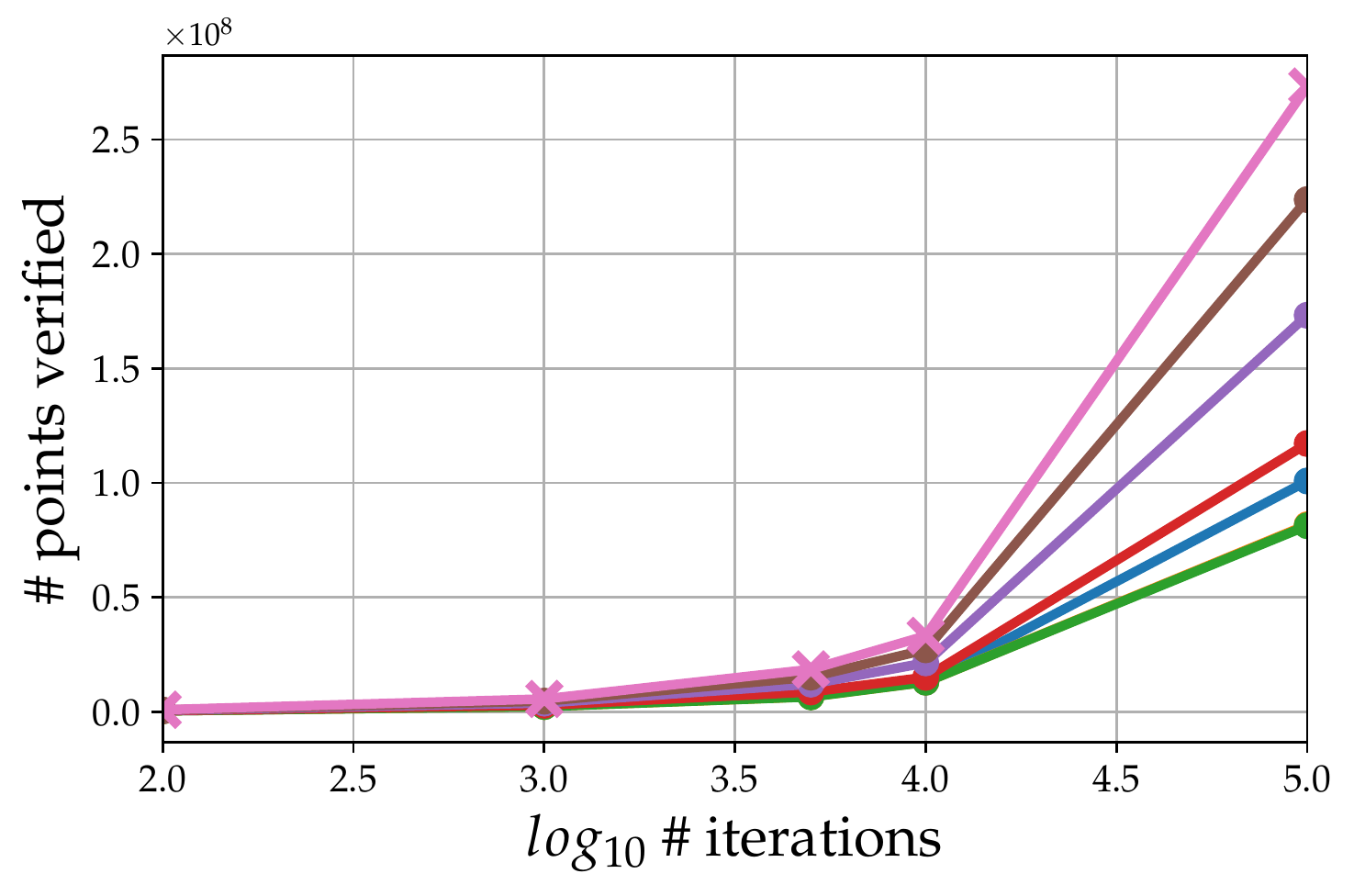}
  	        \includegraphics[width=1.0\columnwidth,trim=0 0 0cm 0cm, clip]{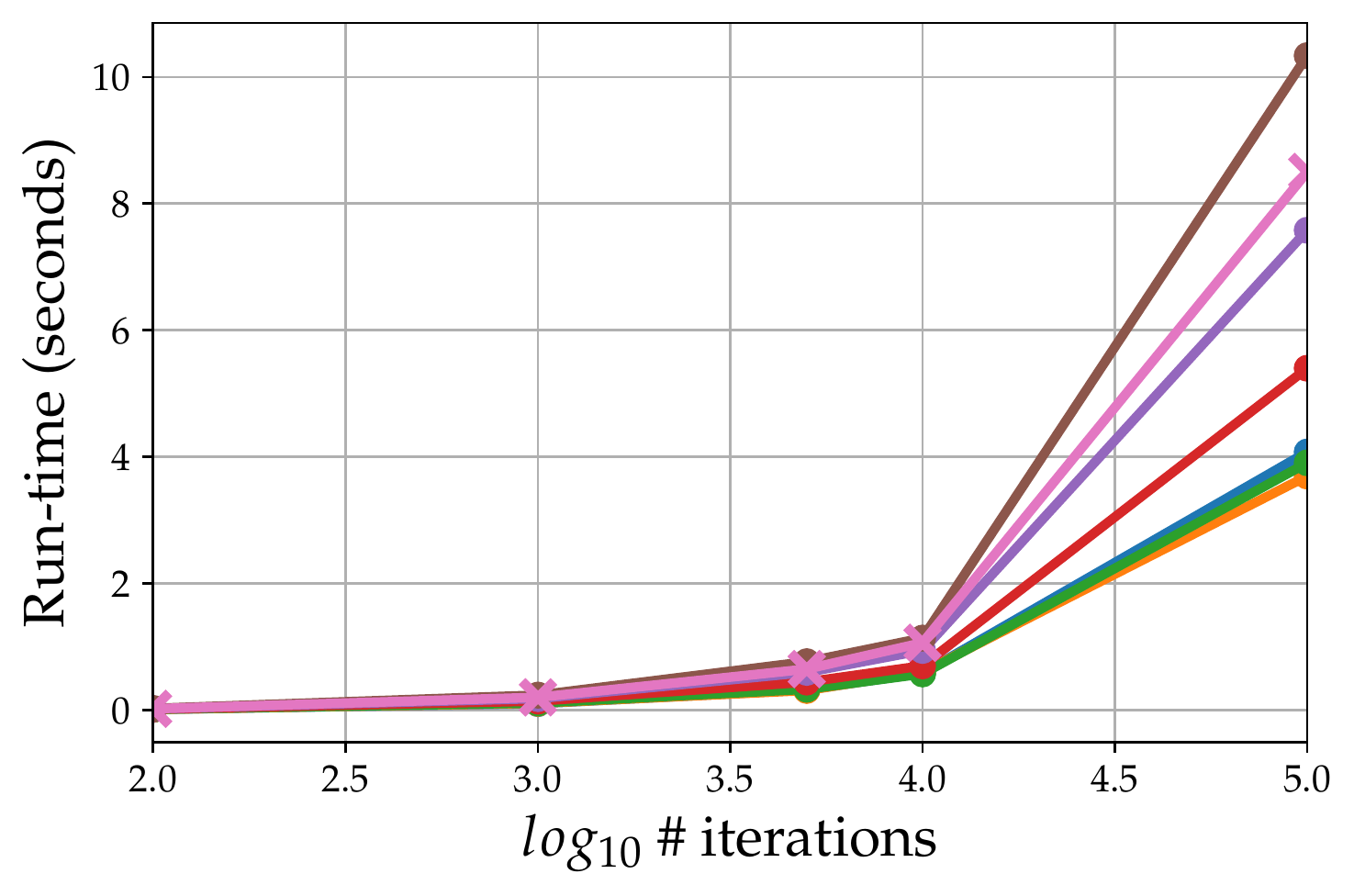}
      		\caption{Essential matrix estimation}
            \label{fig:resultsE}
      \end{subfigure}
      \begin{subfigure}[b]{0.32\linewidth}
     	 	\centering
  	        \includegraphics[width=1.0\columnwidth,trim=0 0 0cm 0cm, clip]{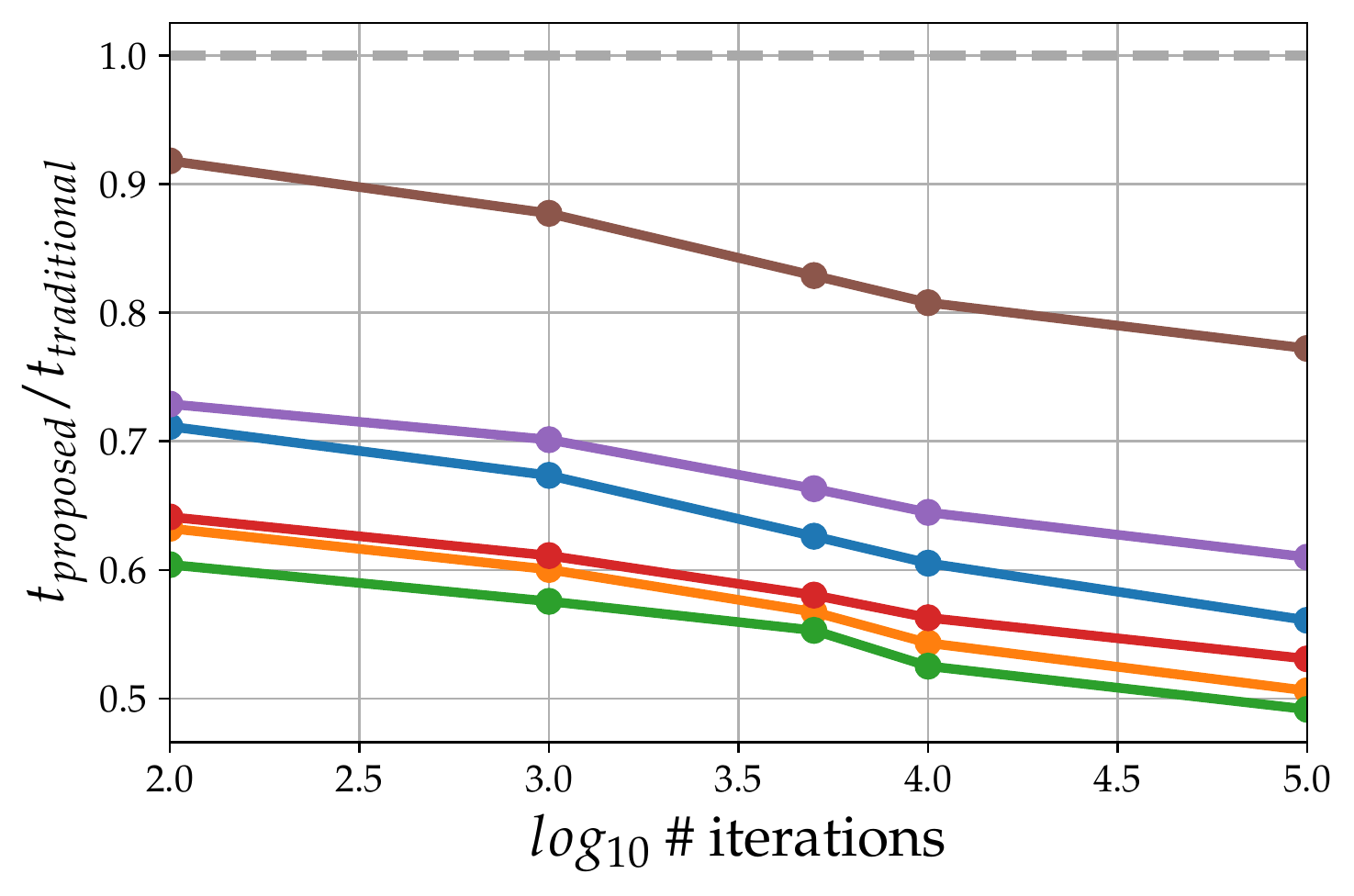}\\
  	        \includegraphics[width=1.0\columnwidth,trim=0 0 0cm 0cm, clip]{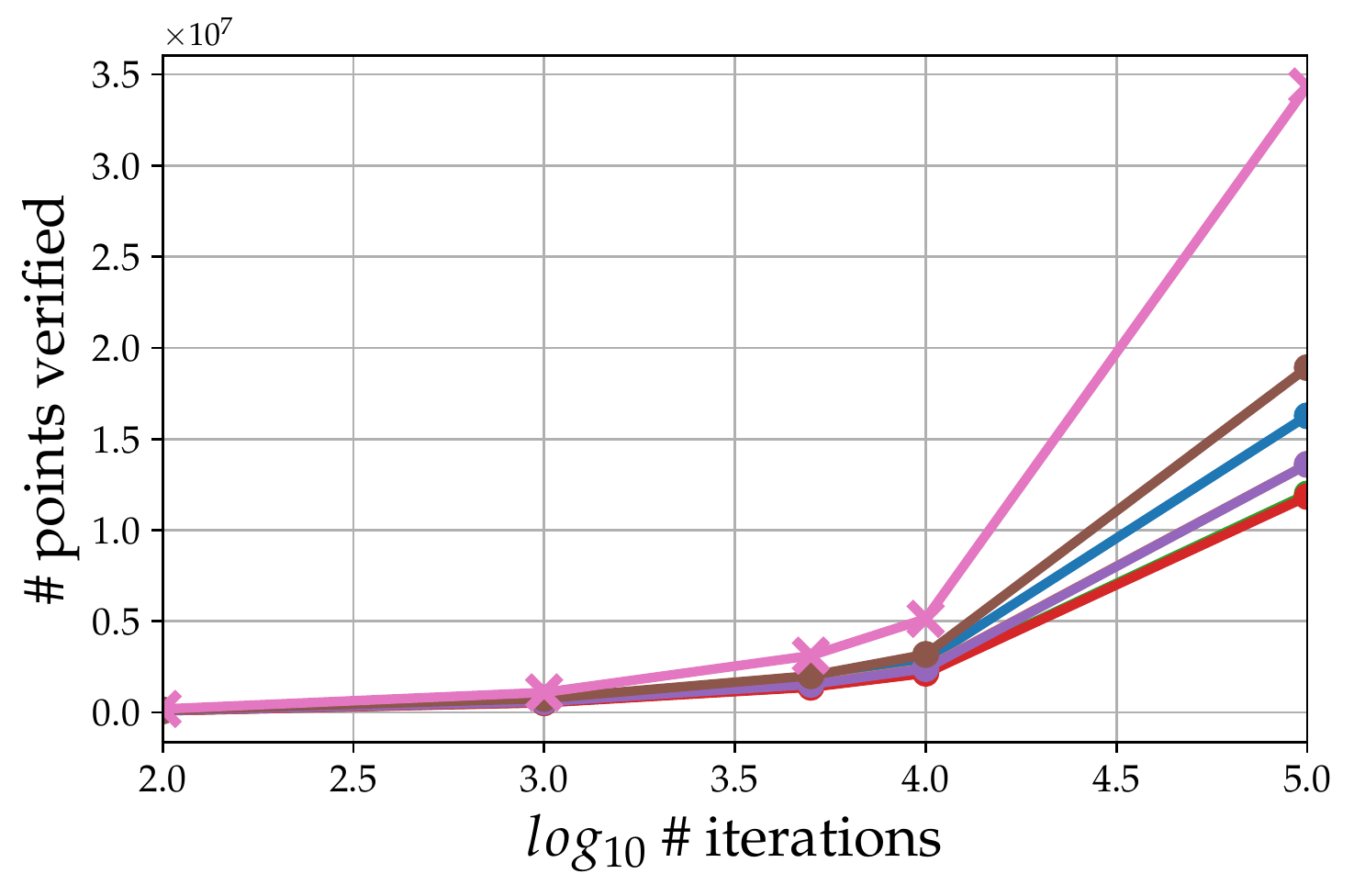}
  	        \includegraphics[width=1.0\columnwidth,trim=0 0 0cm 0cm, clip]{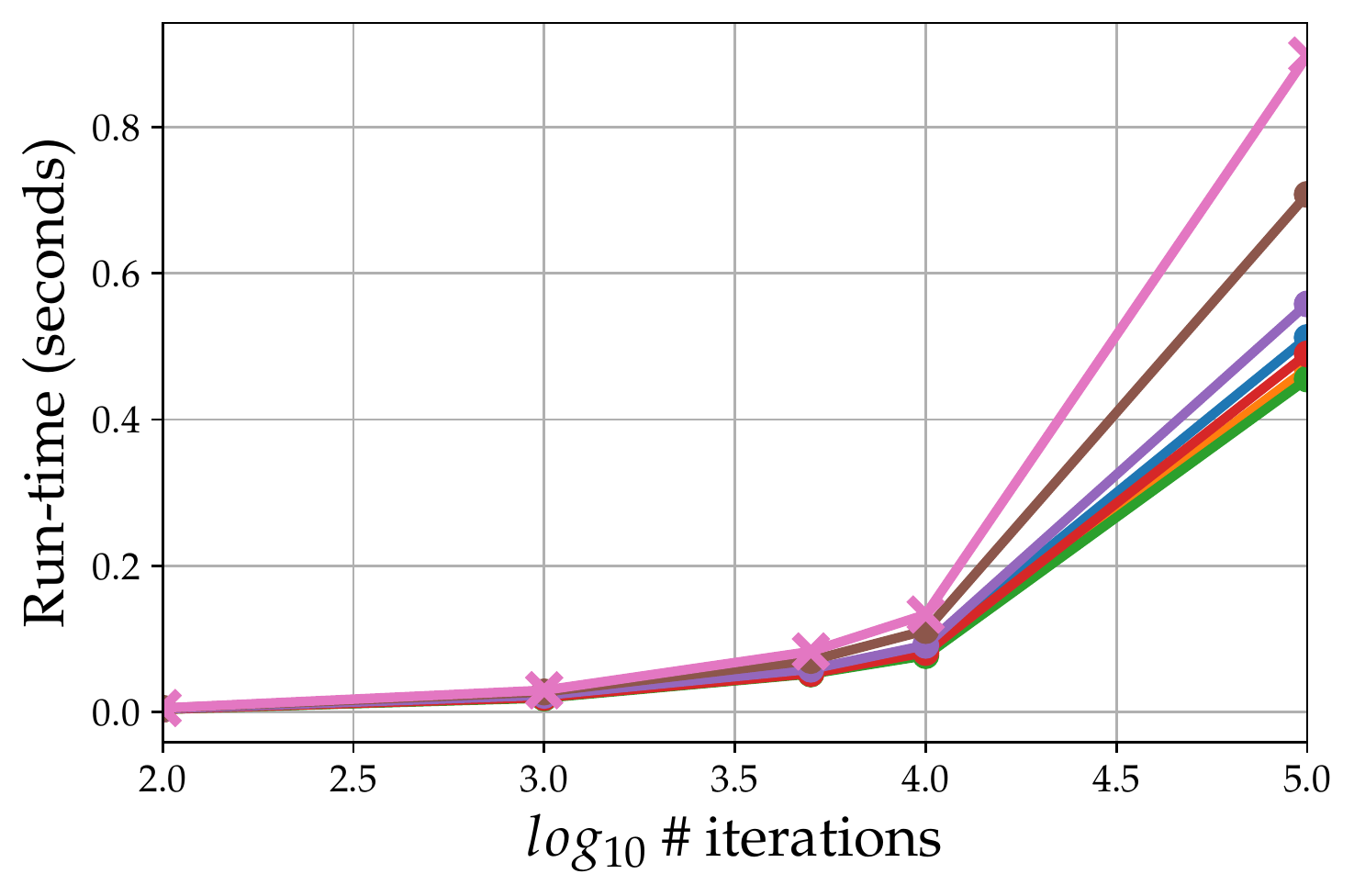}
      		\caption{Homography estimation}
            \label{fig:resultsH}
      \end{subfigure}
  \caption{ The ratio of times of the proposed and traditional methods (top row; vertical axis), the number of points verified within the RANSAC loop (middle) and the run-time in seconds (bottom) are plotted as a function of $\log_{10}$ iteration number (horizontal). The number of 2D-2D cell pairs used in the proposed algorithm is written in brackets.
  For example, if the cell number is $2^4 = 16$, both images were divided into $2$ pieces along each axis.
  We tested $2^4$, $3^4$, $4^4$, $5^4$, $6^4$, and $8^4$ subdivisions running the algorithm on a wide range of cell sizes.
  }
  \label{fig:results_mAA}
\end{figure*}


\begin{figure}[t]
  \centering
  \includegraphics[width=0.32\columnwidth,trim=0 0 0cm 0cm, clip]{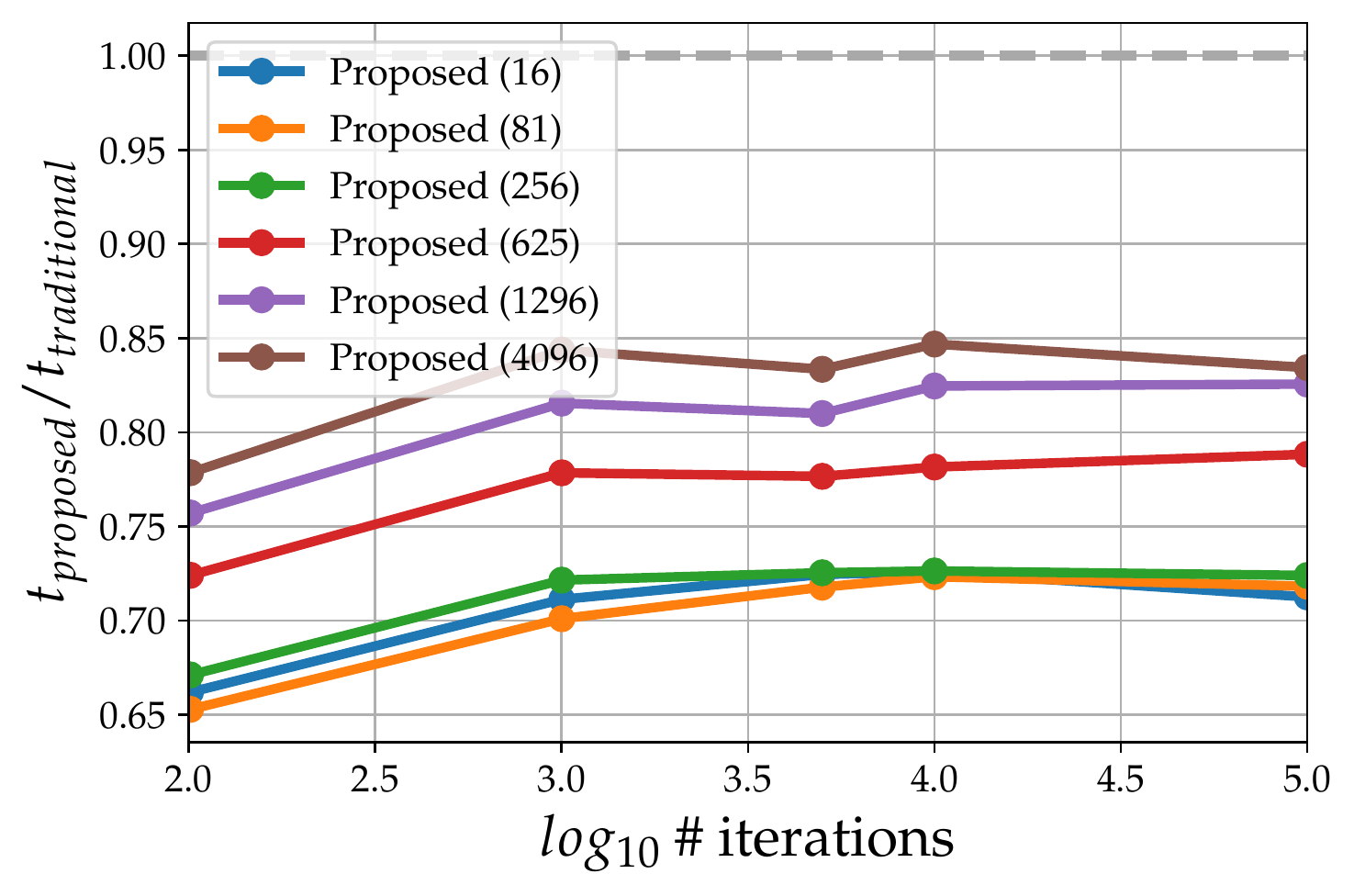}
  \includegraphics[width=0.32\columnwidth,trim=0 0 0cm 0cm, clip]{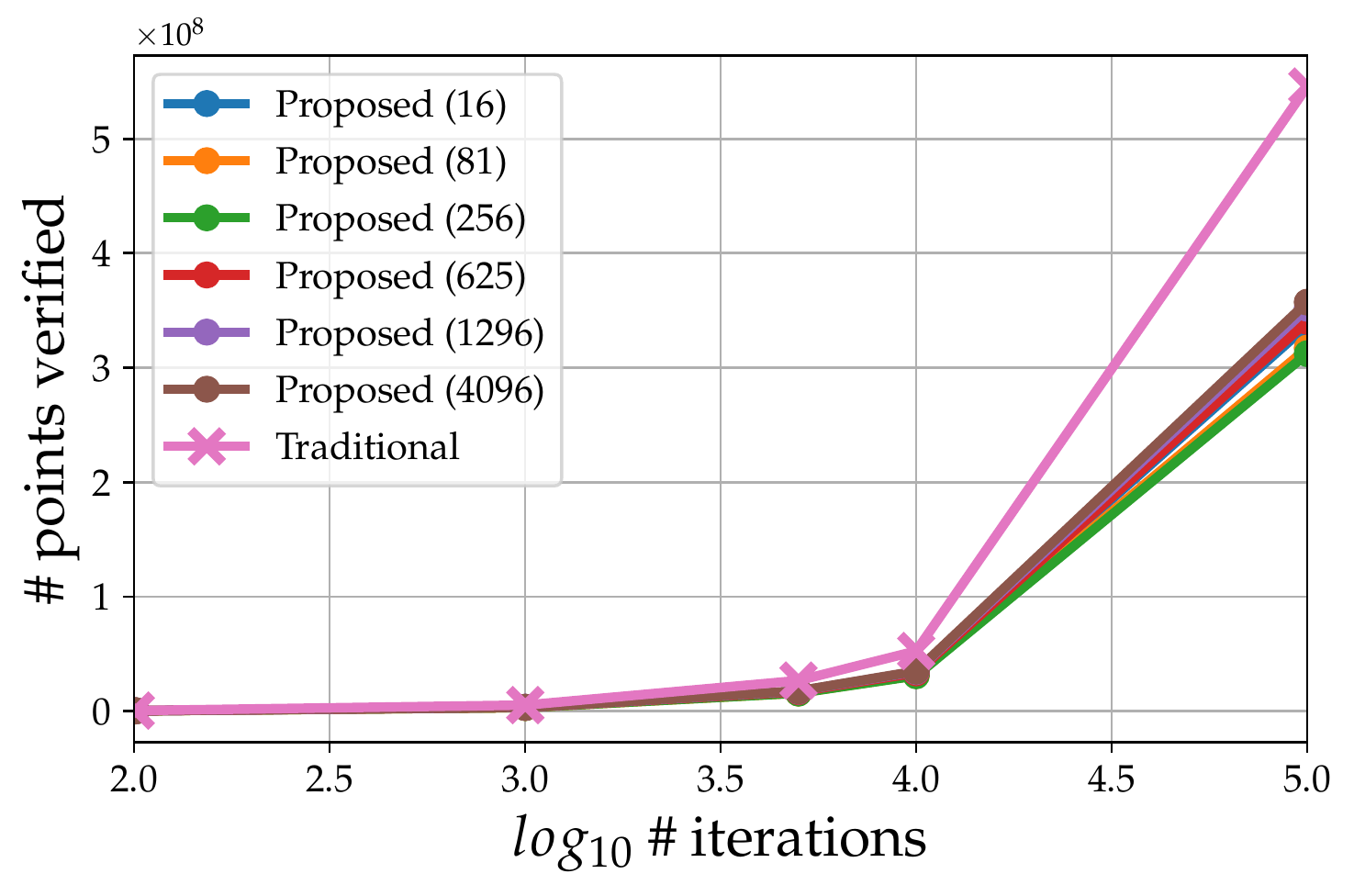}
  \includegraphics[width=0.32\columnwidth,trim=0 0 0cm 0cm, clip]{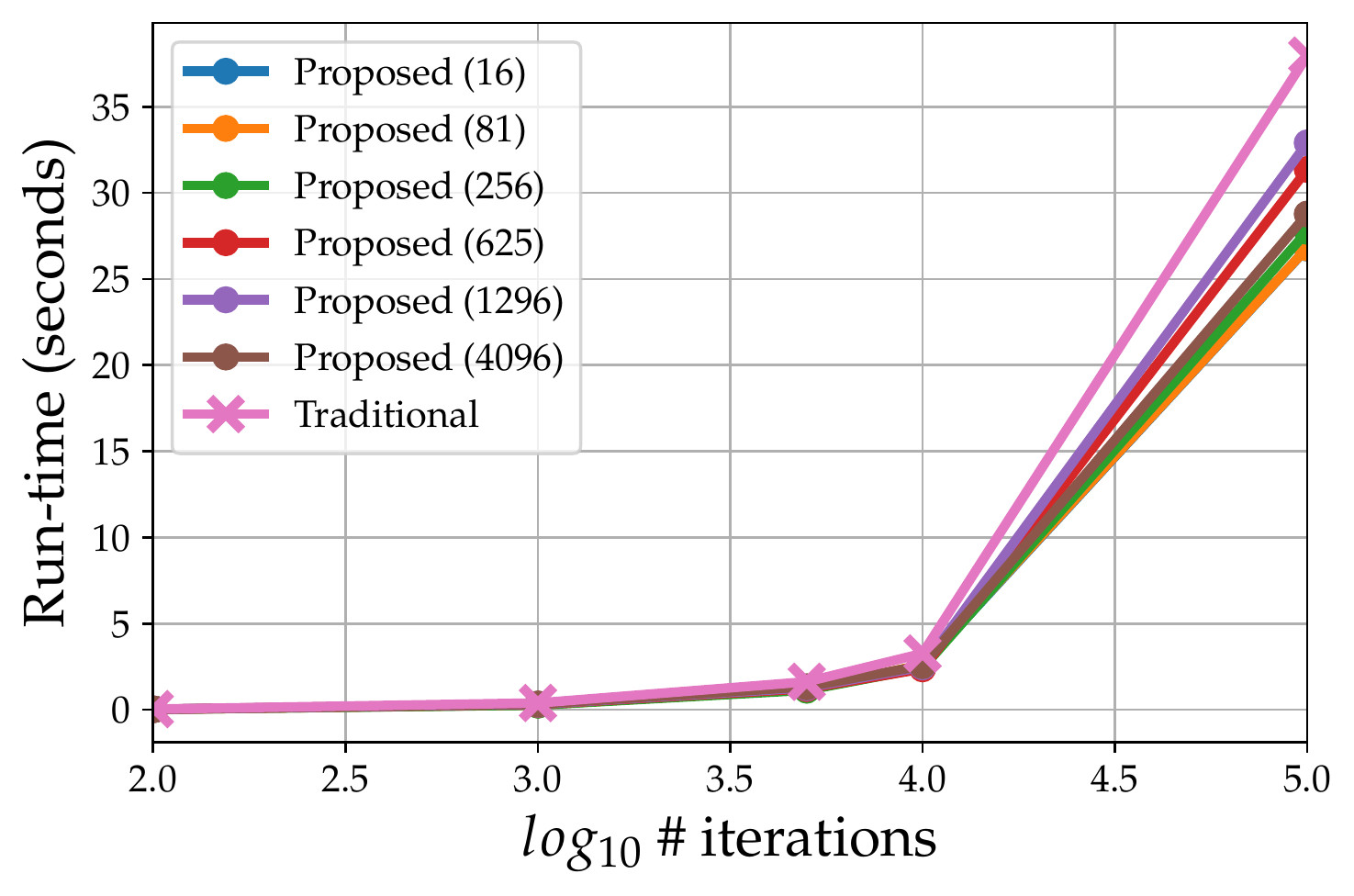}
  \caption{  The ratio of the run-times of the proposed and traditional methods, the number of points verified (middle) and the actual run-time (bottom) are plotted as a function of $\log_{10}$ iteration number for radial homography estimation on 16k image pairs from the Sun360 dataset.}
    \label{fig:resultsRH}
\end{figure}

\noindent \textbf{Homography Estimation.}
For testing the methods on homography estimation, we used the datasets from CVPR tutorial \textit{RANSAC in 2020}~\cite{cvpr2020ransactutorial}.
We used the inlier-outlier threshold for RANSAC tuned in~\cite{cvpr2020ransactutorial}. We ran the method only on scene Sacre Coeur consisting of $4950$ image pairs.  
To form tentative correspondences, we detect $8000$ SIFT keypoints in both images, and use mutual nearest neighbor check with SNN ratio test~\cite{lowe1999object} as suggested in~\cite{IMC2020}.
The average inlier ratio of the tested dataset is 14\% for \textbf{E}/\textbf{F} and 6\% for \textbf{H} estimation ranging from 0.9\% to 68\%. 

In Fig.~\ref{fig:resultsH}, the relative run-time is plotted as the function of the $\log_{10}$ iteration number that was used as a fixed iteration number for RANSAC. 
Early rejection was turned off.
Each curve shows the results of using a regular grid with different number of cells shown in brackets. 
The proposed approach leads to a speed-up with all tested cell numbers.
In this case, $256$ cells (each image is divided into $4 \times 4$ cells) lead to the fastest procedure.
The run-time drops to its $50\%$ when doing \num{10000} iterations.

\begin{figure}[t]
    \centering
      \begin{subfigure}[b]{0.45\columnwidth}
            \includegraphics[width=1.0\columnwidth,trim=0 0 0cm 0cm, clip]{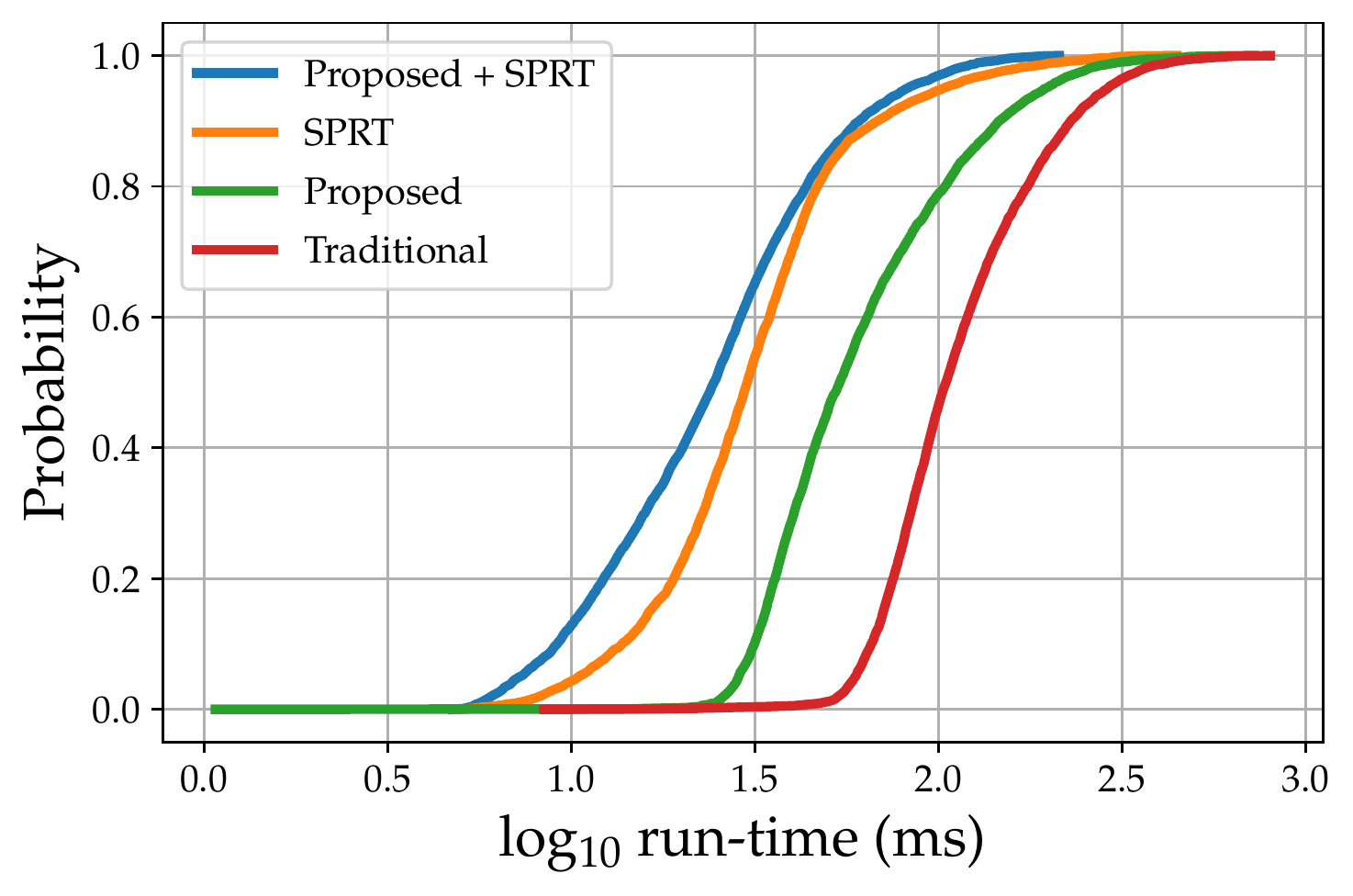}
            \caption{Homography Verification}
            \label{fig:results_cdfH}
      \end{subfigure}
      \begin{subfigure}[b]{0.45\columnwidth}
            \includegraphics[width=1.0\columnwidth,trim=0 0 0cm 0cm, clip]{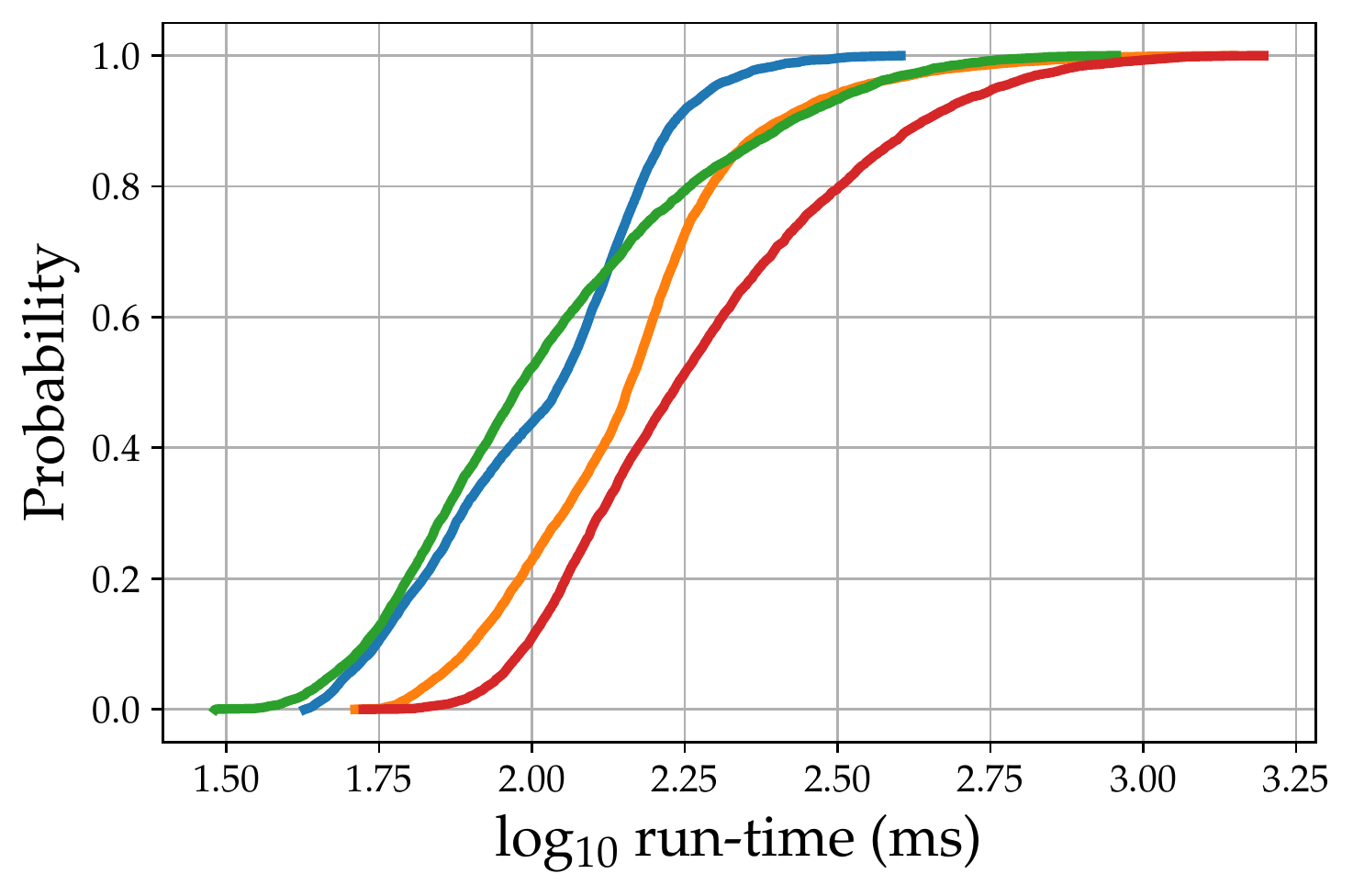}
            \caption{Fundamental Matrix Verification}
            \label{fig:results_cdfF}
      \end{subfigure}\\
      \begin{subfigure}[b]{0.45\columnwidth}
            \includegraphics[width=1.0\columnwidth,trim=0 0 0cm 0cm, clip]{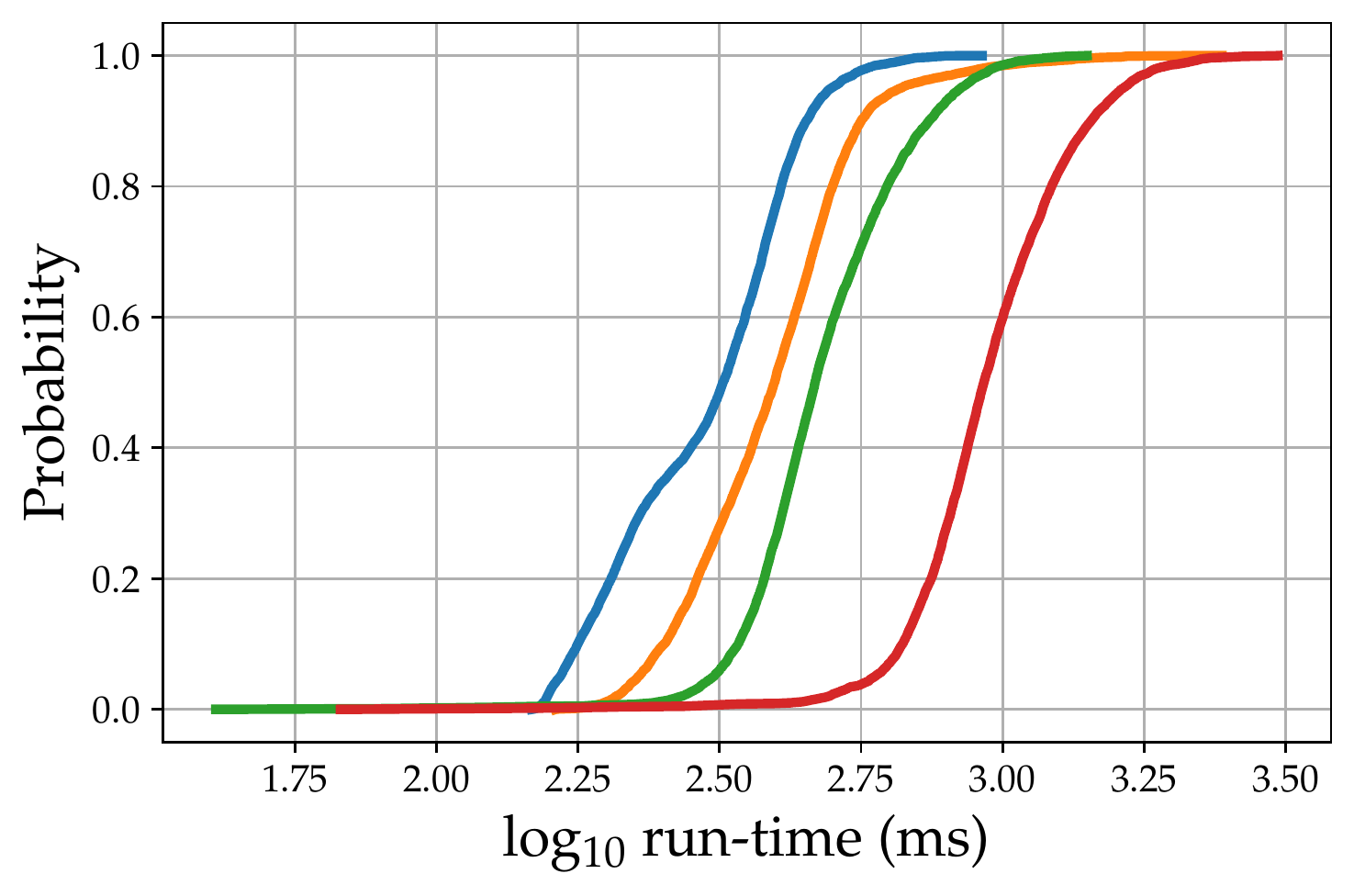}
            \caption{Essential Matrix Verification}
            \label{fig:results_cdfE}
      \end{subfigure}
      \begin{subfigure}[b]{0.45\columnwidth}
            \includegraphics[width=1.0\columnwidth,trim=0 0 0cm 0cm, clip]{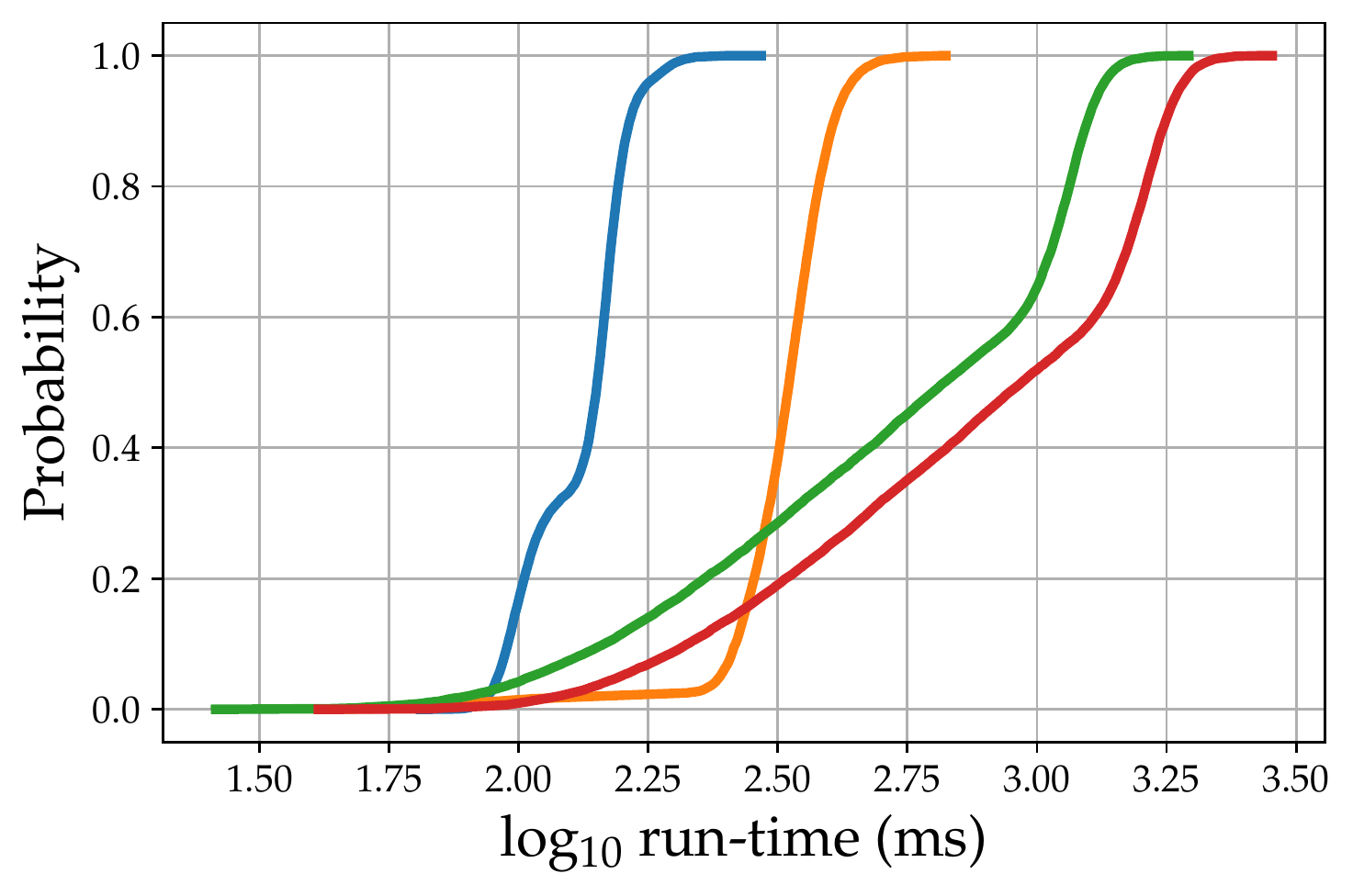}
            \caption{Radial Homography Verification}
            \label{fig:results_cdfR}
      \end{subfigure}
    \caption{ Cumulative distribution functions (CDF) of the times (ms) of the proposed and traditional algorithms, SPRT~\cite{chum2008optimal} and that of SPRT combined with the proposed method on homography, fundamental and essential matrix estimation on 4950 image pairs, and radial homography estimation on 16056 pairs. Being fast is indicated by a curve close to the top-left corner. }
    \label{fig:results_cdf}
\end{figure}

Fig.~\ref{fig:results_cdfH} shows the cumulative distribution functions (CDF) of the run-times (in ms) of SPRT, the proposed and traditional algorithms, and the proposed method with SPRT.
For these experiments, we ran RANSAC with its confidence set to $0.99$ and max.\ iteration number to \num{5000}.
This max.\ iteration number is a strict upper bound, preventing RANSAC to run longer.
We set early rejection threshold $\epsilon_r$ to $1.6$. 
The proposed technique with SPRT runs, on avg.\, for $31.2$ ms, while the avg.\ time of SPRT is $39.3$ ms.

\noindent \textbf{Fundamental Matrix Estimation.}
Same data is used as before.
In Fig.~\ref{fig:resultsF}, the relative run-time (\ie, ratio of the time of the proposed and traditional approaches), the number of points verified in total and the processing time (in seconds) are plotted as the function of the $\log_{10}$ iteration number -- used as a fixed iteration number for RANSAC. 
The proposed early rejection was turned off.
Each curve shows the results of using a regular grid with different number of cells shown in brackets. 
For example, $16 (= 2^4)$ means that each image axis is divided into $2$ parts, thus, having $16$ 2D-2D cells in total. 
For \textbf{F} estimation, 16 cells lead to the fastest calculation with almost halving the run-time of the traditional approach.
The time increases proportionally with the cell number.
This is due to the fact that while having more cells provides a tighter approximation of the inlier set, it requires bounding more cells increasing the problem complexity. 
It is important to note that, similarly as for homography estimation, doing more RANSAC iterations and, thus, likely increasing the accuracy becomes cheaper with the proposed method. 

Fig.~\ref{fig:results_cdfF} shows the cumulative distribution functions (CDF) of the processing times (in milliseconds) of SPRT, the proposed and traditional algorithms, and the proposed method with SPRT.
We set the early rejection threshold $\epsilon_r$ to $1.2$.
The proposed approach with SPRT is the fastest by halving the run-time of the traditional approach and being, on average, faster by 50 ms than SPRT.

\noindent \textbf{Essential Matrix Estimation.}
For essential matrix estimation, we used the same data as for fundamental matrices. 
In Fig.~\ref{fig:resultsE}, the relative run-time, the number of verified points and the actual run-time are plotted as the function of the $\log_{10}$ iteration number that was used as a fixed iteration number for RANSAC. 
Early rejection was turned off.
Each curve shows the results of using a regular grid with different number of cells. 
Similarly as for fundamental matrix estimation, 16 cells lead to the fastest quality calculation.
The speed-up is now even bigger: the run-time drops to its $40\%$ when doing \num{10000} iterations. 
This is caused by the fact that the five-point solver returns a maximum of 10 candidate solutions -- the same number of RANSAC iterations requires more models to be verified than for fundamental matrix estimation.

Fig.~\ref{fig:results_cdfE} shows the cumulative distribution functions (CDF) of the processing times (in milliseconds) of SPRT, the proposed and traditional algorithms, and the proposed method with SPRT.
We set early rejection threshold $\epsilon_r$ to $1.2$. 
It can be seen that the proposed approach causes a quite significant speed-up. 
The proposed technique with SPRT runs, on average, for $317.1$ ms, while the average time of SPRT is $416.2$ ms.

\begin{figure}[t]
  \centering
    \includegraphics[width=0.50\columnwidth,trim=0 0 0cm 0cm, clip]{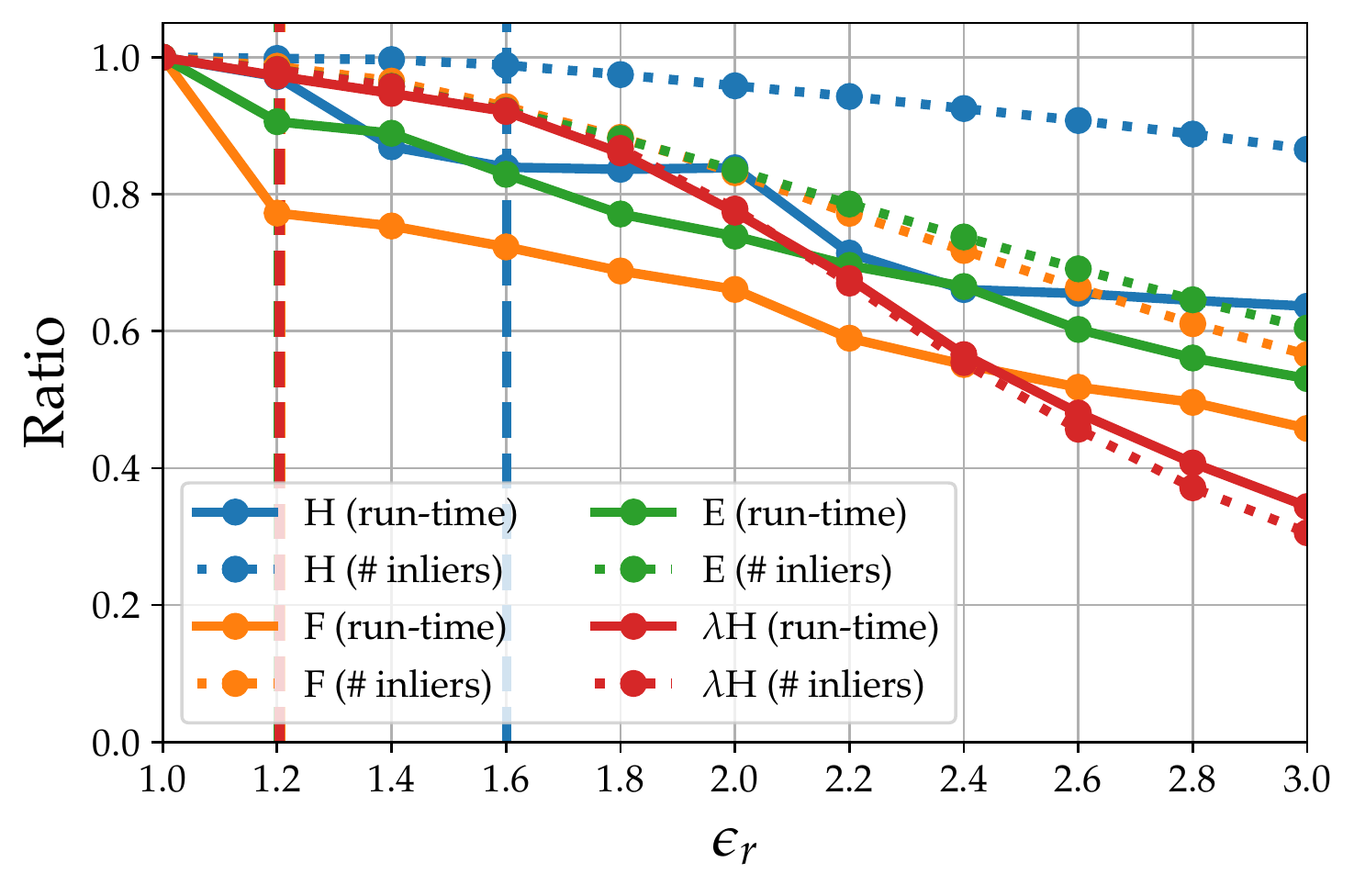}\quad
    \includegraphics[width=0.35\columnwidth]{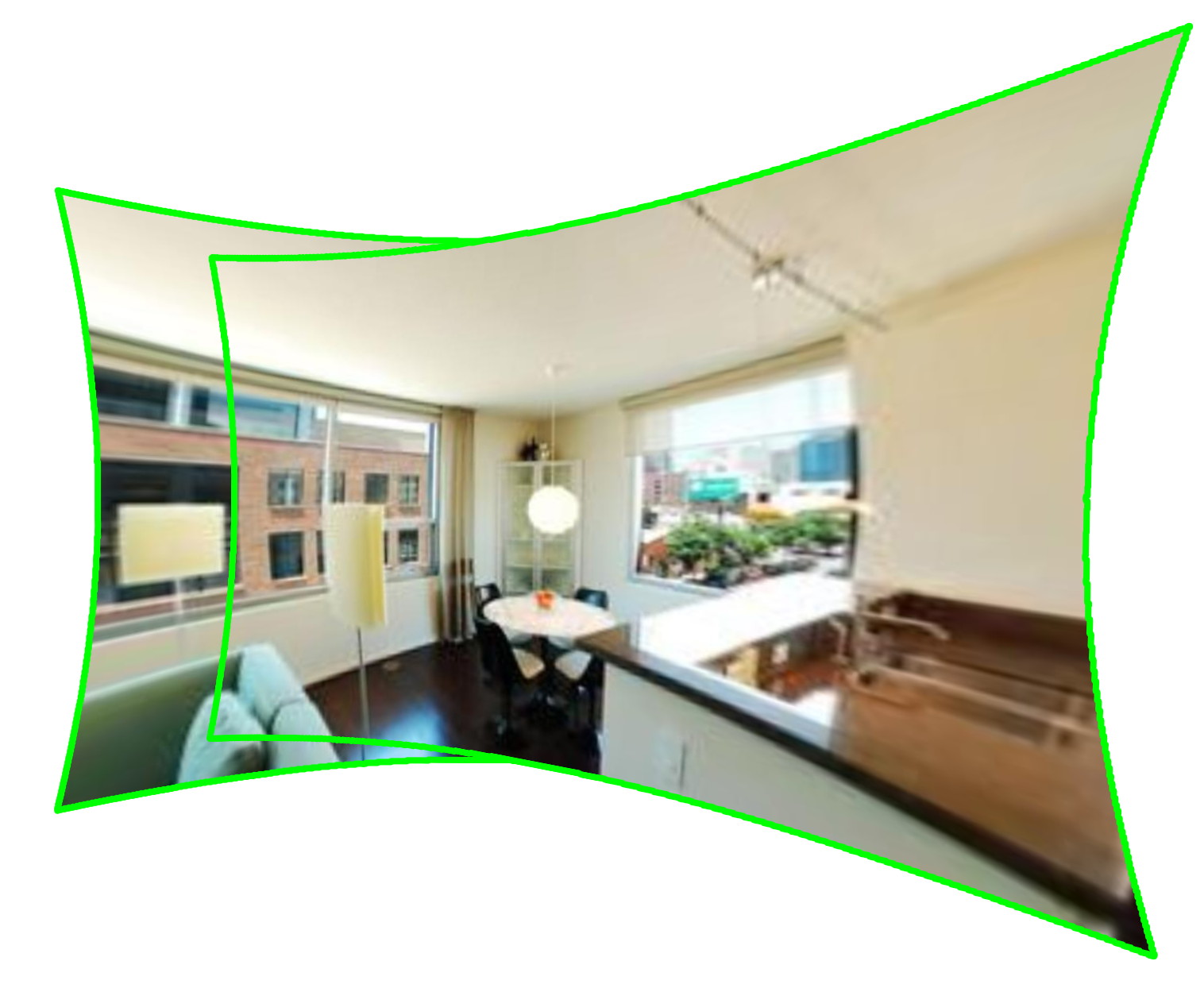}
   \caption{\textbf{(Left)} Cumulative distribution functions (CDF) of the times (in ms) of the proposed and traditional techniques on homography (H), fundamental (F) and essential matrix (E) estimation on 4950 image pairs, and radial homography estimation ($\lambda$H) on 16056 pairs. Being fast is indicated by a curve close to the top-left corner.
    \textbf{(Right)} Radial homography in the Sun360 dataset~\cite{xiao2012recognizing}.}
   \label{fig:stitching_example}
\end{figure}

\noindent \textbf{Radial Homography Estimation.}
To test the proposed techniques on real-world data, we chose the Sun360~\cite{xiao2012recognizing} panorama dataset. 
The purpose of the Sun360 database is to provide academic researchers a comprehensive collection of annotated panoramas covering $360\times180$-degree full view for a large variety of environmental scenes, places and the objects within. To build the core of the dataset, high-resolution panorama images were downloaded and grouped into different place categories. 
To obtain radially distorted image pairs from each 360$^\circ$ panoramic scene, we cut out images simulating a 80$^\circ$ FOV camera with a step size of 10$^\circ$ as done in~\cite{Ding_2021_ICCV}. Thus, the rotation around the vertical axis between two consecutive images is always 10$^\circ$.
Finally, image pairs were formed by pairing the consecutive images in each scene. 
In total, 16056 image pairs were generated. 
For estimating radial distortion homographies from minimal samples, we use the solvers from~\cite{kukelova2015radial}. See Fig.~\ref{fig:stitching_example} for an example image stitching results using a radial homography on an image pair from the Sun360 dataset.

The effect of the grid density is shown in Fig.~\ref{fig:resultsRH}.
The proposed approach accelerates the robust radial homography estimation on the tested wide range of cell numbers.
The best run-times are achieved by partitioning the images into 3 pieces along each axis and, thus, having 81 cell correspondences in total. 

Fig.~\ref{fig:results_cdfR} shows the cumulative distribution functions (CDF) of the processing times (in milliseconds) of SPRT, the proposed and traditional algorithms, and the proposed method with SPRT.
We set early rejection threshold $\epsilon_r$ to $1.2$. 
The proposed technique with SPRT runs, on average, for $134.7$ ms, while the average time of SPRT is $331.3$ ms which is almost three times higher than when using the proposed space partitioning.

\noindent \textbf{Timing Breakdown.}
We show the times spent on each steps of the robust estimation with and without SPRT when using the proposed space partitioning-based verification. 
The times and, also, the accuracy are shown in Table~\ref{tab:time} on homography, essential and fundamental matrix estimation.
The same datasets are used as in the previous sections.
The cell rejection $t_r$ has negligible time demand compared to the verification $t_v$. 
The verification time, when using the proposed approach, is significantly reduced. 
The AUC@$10$ scores are the same without SPRT and similar with SPRT.

\setlength{\tabcolsep}{4pt}
\begin{table}
    \centering
    \begin{tabular}{ c | c | c c | c | c c }
        \hline   
        Problem & SPRT & $t_{r}$ & $t_{v}$ & $t_{v}^{trad}$ & AUC@$10$ & AUC@$10^{trad}$ \\
        \hline   
        \multirow{2}{*}{\textbf{H}} & no & 0.6 & \phantom{1}40.0 & \phantom{1}88.8 & 0.54 & 0.54 \\ 
        & yes & 0.4 & \phantom{11}3.8 & \phantom{11}6.8 & 0.53 & 0.53 \\
        \hline   
        \multirow{2}{*}{\textbf{F}} & no & 0.9 & \phantom{1}50.3 & 113.7 & 0.39 & 0.39  \\
        & yes & 1.1 & \phantom{1}22.5 & \phantom{1}53.2 & 0.38 & 0.37 \\
        \hline   
        \multirow{2}{*}{\textbf{E}} & no & 5.4 & 251.2 & 537.6 & 0.65 & 0.65 \\
        & yes & 4.8 & \phantom{1}29.4 & \phantom{1}79.7 & 0.63 & 0.61 \\
        \hline   
    \end{tabular}
    \caption{The avg.\ (over all image pairs) time spent on cell rejection ($t_r$), model verification using the kept cells ($t_v$), in the traditional verification ($t_{v}^{trad}$) in \textit{ms}; and the AUC@$10$ score of the max.\ of the rotation and translation errors, decomposed from homographies (\textbf{H}) fundamental (\textbf{F}) and essential matrices (\textbf{E}), when using the proposed and traditional approaches.}
    \label{tab:time}
\end{table}

\noindent \textbf{Early Rejection.}
In the left plot of Fig.~\ref{fig:stitching_example}, the change in the run-time and the final inlier number is plotted as the function of the early rejection threshold $\epsilon_r$. 
The results are divided by the result of the $\epsilon_r = 1$ case that provably does not lead to deterioration in the accuracy. 
The vertical lines are placed so the rejection threshold leads to lower than $1\%$ drop in the inlier number.
The green, orange and red lines overlap.
For homographies, $\epsilon_r = 1.6$ leads to negligible accuracy drop while further decreasing the run-time by approximately $20\%$.
For all other tested problems, setting $\epsilon_r$ to $1.2$ is a reasonable choice decreasing the processing time by $22\%$ and $9\%$, respectively.

\section{Conclusion}

We propose a new general algorithm for accelerating the RANSAC model quality calculation.
The method is based on partitioning the joint correspondence space to a pair of regular grids.
Cells of the grids are then projected by each minimal sample model, before calculating its quality, to efficiently reject all correspondences that are inconsistent with the model.   
Besides speeding up the quality calculation significantly, it also allows us to reject models early if the upper bound of their inlier number does not exceed the inlier number of the so-far-the-best model.    
We found that dividing the domain, \eg images, into only a few cells is a good trade-off between getting a tight-enough approximation of the inlier set without significantly increasing the problem complexity. 

The proposed technique reduces the RANSAC run-time by 41\% on average on a wide range of problems and datasets.
When it is combined with the SPRT test, it leads to an approximately $3.4$ times speed-up compared to the traditional algorithm and, also, reduces the SPRT time to its $66$\%. 
It can be straightforwardly inserted into any state-of-the-art robust estimator, \eg, VSAC~\cite{ivashechkin2021vsac} or MAGSAC++~\cite{barath2019magsacpp}, to accelerate them, provably, without any negative side-effect.

{ 
\textbf{Acknowledgments:}
This work was supported by the ETH Zurich Postdoctoral Fellowship.
}

%
%
\bibliographystyle{splncs04}
\bibliography{egbib}

\clearpage
\appendix
\section{Polynomial Approximation}

This section gives a brief overview of three classic polynomial approximation schemes that we experimented with to bound the range of nonlinear transformations. The derivation of the results below are detailed in standard books on numerical analysis, we list these here for implementation reference and to make our paper  self-contained. Similarly, we briefly summarize polynomial basis conversion via fitting for the sake of convenience. 

The literature of polynomial approximations is rich, our selection of Taylor, Hermite, and Lagrange interpolation was motivated by ease of implementation and the existence and conciseness of error terms for these polynomials when used in the context of function approximation. 

In terms of convenience, Lagrange interpolation in Bernstein basis is the least obtrusive solution as it only requires the evaluation of the target function. Hermite and Taylor expansions require higher order derivatives, which have to be either computed formally, via automatic differentiation, or numerical differentiation. However, all solutions require the ability to bound the magnitude of certain derivatives, if conservative bounds are to be computed. 


\subsection{Taylor Approximation}

We denote the $n$-dimensional Euclidean space by $\R^n$ and  $\norm{\cdot}_2$ is the Euclidean norm.
The partial derivatives of an $f:\R^2 \rightarrow \R$ function are $ \partial_1 f,  \partial_2 f$ or $f_x, f_y$.
The scalar product of vectors $\bm a, \bm b \in \R^n$ is written as $\langle\bm a, \bm b\rangle = \bm a\transpose \bm b$.

\begin{definition}
\label{def:multi-index}
Let $\balpha = (\alpha_1, \dots, \alpha_n)\in\N^n$ be a multi-index.
Then we define the following operations:
\begin{itemize}
    \item $|\balpha| = \alpha_1 + \alpha_2 + \dots + \alpha_n$
    \item $\balpha! = \alpha_1! \cdot \alpha_2! \cdot \ldots \cdot \alpha_n!$, where $0! = 1$
    \item $\bm x^\balpha = x_1^{\alpha_1} \cdot x_2^{\alpha_2} \cdot \ldots \cdot x_n^{\alpha_n} \qquad (\bx = (x_1, x_2, \dots, x_n) \in \R^n)$
    \item $\partial^\balpha f = \partial_1^{\alpha_1} \partial_2^{\alpha_2} \dots \partial_n^{\alpha_n} f$ \qquad ($f:\R^n\rightarrow\R)$
\end{itemize}
\end{definition}

\begin{definition} Let $f : \mathbb{R}^n \rightarrow \mathbb{R}$ be $f \in C^{k+1}$. The degree $k$ multivariate Taylor 
approximation of $f$ about $\bm x_0$ is
\begin{equation}
\label{eq:Taylor_n}
T_{k, \bm x_0}(\bm x) = \sum_{|\balpha| \leq k} \frac{ \partial^\balpha f(\bm{x}_0) }{\balpha!} ( \bm{x} - \bm{x}_0 )^\balpha ~.
\end{equation}
\end{definition}

\begin{theorem}[Taylor Approximation Theorem]
\label{th:taylor-approx-multivar}
Let $f:\R^n\rightarrow\R, S\subset \R^n$ open and convex, $f\in C^{k+1}[S]$.
If $\bm a, \bm a + \bm h \in S$, then
\begin{equation}
    f( \bm a + \bm h ) = T_f^{(k)}(\bm a + \bm h) + R_{\bm a, k}(\bm h) 
\end{equation}
where the residual $R_{\bm a, k}$ can be expressed using an adequate $c \in (0, 1)$:
\begin{equation}
    R_{\bm a, k}(\bm h) = \sum_{|\balpha|=k+1} \partial^\balpha f( \bm a + c\cdot \bm h) \frac{ \bm h^\balpha }{ \balpha! }
\end{equation}
or, with an integral form, as
\begin{equation}
    R_{\bm a, k}(\bm h) = (k+1) \sum_{|\balpha| = k+1} \frac{ \bm h^\balpha }{ \balpha! } \int_0^1 (1-t)^k \partial^\balpha f(\bm a + t \bm h) dt ~.
\end{equation}
\end{theorem}

\begin{corollary}[Error bound of Taylor approximation] 
Let $f : \R^n \rightarrow \R$ be such that $f \in C^{k+1}[S]$ and $M>0$ such that $\forall \bm x \in S : \forall |\balpha| = k+1 : | \partial^{\balpha}f( \bm x ) | \leq M$. Then
\begin{equation}
    R_{\bm a, k}( \bm h ) \leq \frac{ M }{ (k+1)! } || \bm h ||_1^{k+1} ~.
\end{equation}
\end{corollary}

In the two-dimensional case, the Taylor polynomials are written as
\begin{equation}
T^{(k)}_f( x, y ) = \sum_{i=0}^k\sum_{j=0}^i \frac{\partial_1^j\partial_2^{i-j} f(a, b)}{j!(i-j)!}(x-a)^j(y-b)^{i-j}
\end{equation}

In the case of vector valued functions of two variables, the Taylor expansion naturally generalizes to
\begin{equation}
\bm{T}^{(k)}_{\bm f}( x, y ) = \sum_{i=0}^k\sum_{j=0}^i \frac{\partial_1^j\partial_2^{i-j} \bm{f}(a, b)}{j!(i-j)!}(x-a)^j(y-b)^{i-j} \in \R^n
\end{equation}

\subsection{Hermite Interpolation}

\begin{definition}[Hermite interpolant]
Let $f : \R \rightarrow \R$ be such that $f \in C^{k+1}$. The polynomial  $h_k(x)$ of degree $2k+1$ is an order $k$ Hermite interpolant at $a, b \in \mathbb{R}$ if and only if
\begin{gather}
    h_k^{(i)}(a) = f^{(i)}(a) \\
    h_k^{(i)}(b) = f^{(i)}(b)
\end{gather}
holds, $i=0, \dots, k$ and $f^{(i)}$ denotes the $i$-th derivative.
\end{definition}

The above two-point Hermite interpolation is sometimes described as dense in the sense that all derivatives and function values are prescribed up to a fixed order and there are no gaps, that is, missing derivatives. 
It can be easily seen that the Hermite interpolation polynomial is unique. More importantly, its error characteristics are given by 

\begin{theorem}[Error bound of Hermite interpolation] 
Let $f : \R \rightarrow \R$ be such that $f \in C^{k+1}$ and let $h_k(x)$ be an order $k$ Hermite interpolant at $a, b \in \mathbb{R}$. Then for all $x \in [a, b]$ exists a $\xi \in [a,b]$ such that
\begin{equation}
    f(x) - h_k(x) = \frac{ f^{(k+1) }(\xi) }{(k+1)!} (x - a)^{k+1} (x - b)^{k+1} ~.
\end{equation}
\end{theorem}

Oftentimes, it is more convenient to bound the above as a function of the $b-a$ width of the domain. The maximum of the function is attained at the midpoint of the interval and straightforward substitution gives the resulting modified bound. 
The above holds for vector valued functions as well but similarly to the Taylor case, the 1-norm has to be used. 

In our case, we approximate the image of the cell boundary curves, thus the single variable error term is sufficient. 

\subsection{Lagrange Interpolation}

By Lagrange interpolation we refer to the interpolation of a $f : [a,b] \rightarrow \mathbb{R}, [a,b] \subset \mathbb{R}$ function at some prescribed $a = x_0 < x_1 < \dots < x_k =b$ points by polynomials. Then the following holds
\begin{theorem} If $p_k(x)$ is a polynomial that interpolates $f : [a,b] \rightarrow \mathbb{R}, [a,b]$ at $a = x_0 < x_1 < \dots < x_k =b$ and $f \in C^{k+1}[a,b]$, then for any $x \in [a,b]$, there exists a $\xi \in (a, b)$ such that the following holds:
\begin{equation}
    \label{eq:LagrangeError} f(x) - p_k(x) = (x-x_0)\cdot \dots \cdot(x-x_k) \frac{f^{(k+1)}(\xi)}{(n+1)!} ~.
\end{equation}
\end{theorem}

There are ways to re-phrase the above in terms of differences, should the target function not meet the continuity assumptions of the theorem but we did not experiment with the practical applicability of these.

As we have no control over the magnitude of $f^{(k+1)}(\xi)$, the only way to minimize the error in \eqref{eq:LagrangeError} is to find $x_i \in [a,b]$ nodes that minimize $\Pi_{i=0}^k (x - x_k)$ over $[a,b]$.  

\begin{definition}[Chebyshev polynomials]
The Chebyshev polynomials over $[-1,1]$ are defined recursively as
\begin{gather}
        T_0(x) = 1, \\
    T_1(x) = x, \\
    T_{k+1}(x) = 2x T_k(x) - T_{k-1}(x)
\end{gather}
for $k \geq 1$.
\end{definition}

\begin{theorem}[Interpolation at Chebyshev nodes] If $p_k(x)$ is a polynomial that interpolates $f : [-1,1] \rightarrow \mathbb{R}, f \in C^{k+1}[a,b]$ at the roots of $T_{k+1}(x)$, that is, $x_i = \cos \left( \frac{2i + 1}{2k+2} \right)$, $i=0, 1, \dots, k$, then 
\begin{equation}
    | f(x) - p_k(x) | \leq \frac{1}{2^k (k+1)!} \max_{t\in[-1,1]} \left| f^{(k+1)}(t) \right| ~.
\end{equation}
This is the best upper bound if we can only vary the location of the $x_i$ interpolation nodes.
\end{theorem}

If the function is defined over an arbitrary $[a,b]$ interval, the Chebyshev nodes simply have to be affinely mapped from $[-1,1]$ to $[a,b]$ to compute the necessary Chebyshev nodes as follows:
\begin{equation}
    x_i = \frac{a+b}{2} + \frac{b-a}{2} \cos \left( \frac{2i+1}{2k+2} \right) ~.
\end{equation}

\section{Bounding Polynomials}

\subsection{Properties of B\'ezier Curves}

Let $\bm b_i \in \mathbb{R}^d, (~i=0, 1, \dots, n)$ denote the control points of a $d$-dimensional Bézier curve. The parametric equation of the curve is
\begin{equation}\label{eq:Bezier}
    \bm b(t) = \sum_{i=0}^{n} \bm b_i B_i^n(t) ~,~~ t\in[0,1] ~,
\end{equation}
where $B_i^n(t)$ are the Bernstein polynomials over $[0,1]$, \ie
\begin{equation}
    B_i^n(t) = \binom{n}{i} t^i (1-t)^{n-i} ~.
\end{equation}

As the Bernstein basis is positive and forms a partition of unity (i.e. $B_i^n(t) \geq 0, t \in [0,1]$ and $\sum_{i=0}^n B_i^n(t) = 1$), it follows that all points of the curve are contained within the convex hull of its $\bm b_i$ control points. Consequently, the axis aligned bounding box of the control points is a conservative bound on the range of the curve. 

Similarly, if we want to bound the magnitude of a $\bm f : \mathbb{R} \rightarrow \mathbb{R}^n$ vector valued function, we can construct a B\'ezier approximation to the $\lVert \bm f( x_i ) \rVert$ magnitude values (in arbitrary norm) via interpolation and use the value of the largest control point (here, scalar) to infer an approximate upper bound on the magnitude.

\subsection{Interpolating Data}

Recall that the evaluation of a function in a basis such as in Equation (\ref{eq:Bezier}) can be written in matrix form as
\begin{equation}
    \bm b(t) = [ B_0^n(t), B_1^n(t), \dots, B_n^n(t) ] \cdot \bmat{ \bm b_0 \\ \bm b_1 \\ \hdots \\ \bm b_n }
\end{equation}

As such, when given $n+1$ parameter values $t_0 < t_1 < \dots < t_n$ and corresponding points in space $\bm p_0, \bm p_1, \dots, \bm p_n$, we can compute the $\bm b_i, (i=0, 1, \dots, n)$ B\'ezier control points that interpolate them by solving
\begin{equation}
    \bmat{ 
        B_0^n(t_0) & B_1^n(t) & \dots & B_n^n(t) \\
        B_0^n(t_1) & B_1^n(t) & \dots & B_n^n(t) \\
        \dots & \dots & \dots & \dots \\
        B_0^n(t_n) & B_1^n(t_n) & \dots & B_n^n(t_n)
    } 
    \cdot 
    \bmat{ \bm b_0 \\ \bm b_1 \\ \hdots \\ \bm b_n }
    =
    \bmat{ \bm p_0 \\ \bm p_1 \\ \hdots \\ \bm p_n }
\end{equation}
for $[\bm b_0, \dots, \bm b_n]^T$. One can either use a linear solver for better robustness, or use interpolation nodes that yield a small condition number for the matrix on the left. Chebyshev nodes are such a choice, up to moderate degrees (that is, up to 10), making direct inversion possible which reduces the interpolation problem to a simple matrix-vector multiplication. Note that if $t_0 = 0, t_n = 1$, the first and the last rows of the matrix are $\bm e_1, \bm e_{n+1}$ respectively, where $\bm e_i$ are the canonical basis vectors of dimension $n+1$. 

In our tests on Lagrange interpolation, we used the roots of the Chebyshev polynomials over closed intervals, i.e. $t_i = \frac{1}{2} \left( 1 - \cos(\frac{ i \pi }{n}) \right) \in [0,1], i=0, \dots, n$. 

\subsection{Converting Hermite to B\'ezier Control Data}

Since we approximate our mapped boundary curves from endpoint derivative data, \ie we use Hermite interpolation, we have to convert the Hermite basis polynomial data to Bernstein basis. This can be done by brute-force interpolation, as in evaluating the Hermite polynomial in $n+1$ points and multiplying the resulting vector by the inverse of the Bernstein evaluation matrix of at the sample parameters, as shown in the previous subsection. 

A simpler approach is possible, however, by recalling that the derivatives of Bézier curves at the endpoints are
\begin{align}
    \bm b^{(k)}(0) &= \frac{n!}{(n-k)!} \Delta^k \bm b_0 \\
    \bm b^{(k)}(1) &= \frac{n!}{(n-k)!} \Delta^k \bm b_{n-k}
\end{align}
where the $\Delta$ forward differences are defined as
\begin{equation}
    \Delta^j \bm b_i = \Delta^{j-1} \bm b_{i+1} - \Delta^{j-1} \bm b_i 
\end{equation}
for $j=1, 2, \dots$ and $\Delta^0 \bm b_i = \bm b_{i+1} - \bm b_i$.
These allow us to compute the control points directly from the raw derivatives. 

Let $\bm m_{i}^{(k)}, i=0,1$ denote the appropriate $k$-th directional derivatives at the two endpoints of the boundary curve. Then from requiring
\begin{equation}
    \bm m_{i}^{(k)} = \bm b^{(k)}(i) \quad,~ (i = 0, 1)
\end{equation}
to hold, we have
\begin{align}
    \bm m_{0}^{(k)} &= \frac{n!}{(n-k)!} \Delta^k \bm b_0 \\
    \bm m_{1}^{(k)} &= \frac{n!}{(n-k)!} \Delta^k \bm b_{n-k}
\end{align}
This allows us to progressively compute the control points from the derivatives such that the resulting curve will reconstruct them at the endpoints. 

For example, the first three derivatives at $t=0$ determine the $\bm b_1, \bm b_2, \bm b_3$ control points from
\begin{align}
    \bm m_0^{(1)} &= n ( \bm b_1 - \bm b_0 ) \\
    \bm m_0^{(2)} &= n (n-1) ( \bm b_2 - \bm b_1 - \Delta \bm b_0 ) \\
    \bm m_0^{(3)} &= n (n-1) (n-2) ( \bm b_3 - \bm b_2 - \Delta \bm b_1 - \Delta^2 \bm b_0 )
\end{align}
as
\begin{align}
    \bm b_1 &= \frac{\bm m_0^{(1)}}{n} + \bm b_0 \\
    \bm b_2 &= \frac{\bm m_0^{(2)}}{n(n-1)} + \Delta \bm b_0 + \bm b_1 \\
    \bm b_3 &= \frac{\bm m_0^{(3)}}{n(n-1)(n-2)} + \Delta^2 \bm b_0 + \Delta \bm b_1 + \bm b_2
\end{align}
and at $t=1$ endpoint from
\begin{align}
    \bm m_1^{(1)} &= n ( \bm b_{n} - \bm b_{n-1} ) \\
    \bm m_1^{(2)} &= n (n-1) ( \Delta \bm b_{n-1} - \bm b_{n-1} + \bm b_{n-2} ) \\
    \bm m_1^{(3)} &= n (n-1) (n-2) ( \Delta^2 \bm b_{n-2} - \Delta^2 \bm b_{n-3} ) \\
\nonumber    &= n (n-1) (n-2) ( \Delta^2 \bm b_{n-2} - \Delta \bm b_{n-2} + \bm b_{n-2} - \bm b_{n-3} )    
\end{align}
as
\begin{align}
    \bm b_{n-1} &= \bm b_{n} - \frac{ \bm m_{1}^{(1)} }{n}  \\
    \bm b_{n-2} &= \bm b_{n-1} - \Delta \bm b_{n-1} + \frac{ \bm m_{1}^{(2)} }{n(n-1)}  \\
    \bm b_{n-3} &= \bm b_{n-2} - \Delta \bm b_{n-2} + \Delta^2 \bm b_{n-2} - \frac{ \bm m_{1}^{(3)} }{n(n-1)(n-2)}
\end{align}

\end{document}